\documentclass{article}
\newcommand{\shorttitle}{}
\DeclareUnicodeCharacter{2264}{\leq}
\DeclareUnicodeCharacter{202F}{\,}

\usepackage{arxiv}
\usepackage[utf8]{inputenc}
\usepackage[T1]{fontenc}
\usepackage{url}
\usepackage{booktabs}
\usepackage{amsmath,amssymb,amsfonts}
\usepackage{nicefrac}
\usepackage{graphicx}
\graphicspath{ {./images/} }
\usepackage{pifont}
\usepackage{dsfont}
\usepackage{float}
\usepackage{longtable}
\usepackage{cite}
\usepackage{textcomp}
\usepackage{listings}
\usepackage{xcolor}
\usepackage{multirow}
\usepackage{subcaption}
\usepackage{colortbl}
\usepackage{array}
\usepackage{bbm}
\usepackage{hyperref}       % Second to last
\usepackage{cleveref}       % LAST

\newcommand{\cmark}{\ding{51}}
\newcommand{\xmark}{\ding{55}}

\def\BibTeX{{\rm B\kern-.05em{\sc i\kern-.025em b}\kern-.08em
    T\kern-.1667em\lower.7ex\hbox{E}\kern-.125emX}}

\newcommand{\QSTD}{\textsc{QSTD}}

\newcommand{\INCORE}{\textsc{IN-CORE}}

\title{Architectural Insights for Post-Tornado Damage Recognition}

\author{
    Robinson Umeike$^{1}$\thanks{Corresponding author. Email: \texttt{crumeike@crimson.ua.edu}} , 
    Thang Dao$^{1}$, 
    Shane Crawford$^{1}$, 
    John van de Lindt$^{2}$, 
    Blythe Johnston$^{3}$, \\
    Wanting (Lisa) Wang$^{4}$, 
    Trung Do$^{5}$, 
    Ajibola Mofikoya$^{1}$, 
    Sarbesh Banjara$^{1}$, 
    Cuong Pham$^{5}$ \\
    \\
    $^{1}$The University of Alabama \quad
    $^{2}$Colorado State University \quad
    $^{3}$Argonne National Laboratory \\
    $^{4}$Old Dominion University \quad
    $^{5}$University of South Alabama \\
}

\begin{document}
\maketitle
\begin{abstract}
Rapid and accurate building damage assessment in the immediate aftermath of tornadoes is critical for coordinating life-saving search and rescue operations, optimizing emergency resource allocation, and accelerating community recovery. However, current automated methods struggle with the unique visual complexity of tornado-induced wreckage, primarily due to severe domain shift from standard pre-training datasets and extreme class imbalance in real-world disaster data. To address these challenges, we introduce a systematic experimental framework evaluating 79 open-source deep learning models, encompassing both Convolutional Neural Networks (CNNs) and Vision Transformers, across over 2,300 controlled experiments on our newly curated Quad-State Tornado Damage (\QSTD{}) benchmark dataset. Our findings reveal that achieving operational-grade performance hinges on a complex interaction between architecture and optimization, rather than architectural selection alone. Most strikingly, we demonstrate that optimizer choice can be more consequential than architecture: switching from Adam to SGD provided dramatic F1 gains of +25 to +38 points for Vision Transformer and Swin Transformer families, fundamentally reversing their ranking from bottom-tier to competitive with top-performing CNNs. Furthermore, a low learning rate of $1\times10^{-4}$ proved universally critical, boosting average F1 performance by +10.2 points across all architectures. Our champion model, ConvNeXt-Base trained with these optimized settings, demonstrated strong cross-event generalization on the held-out Tuscaloosa-Moore Tornado Damage (TMTD) dataset, achieving 46.4\% Macro F1 (+34.6 points over baseline) and retaining 85.5\% Ordinal Top-1 Accuracy despite temporal and sensor domain shifts. This work establishes a novel large-scale benchmark for automated tornado damage assessment and provides actionable optimization guidelines for deploying robust, open-source assessment systems.
\end{abstract}

% keywords can be removed

\section{Introduction}
Tornadoes represent one of the most destructive and sudden natural hazards. The rapid assessment of building damage following these events is critical for emergency response coordination, resource allocation, and recovery planning \cite{b1, b2, b3, b4}. This urgency has driven a shift away from slow, traditional assessment methods toward automated solutions \cite{b5, b6, b7, b8}.
Historically, damage assessment relied on subjective, time-intensive manual inspections by ground teams. While rigorous engineering taxonomies like the IN-CORE framework exist to standardize evaluations based on component failures (e.g., roof covering, windows, structural connections) \cite{b9}, their manual application is a critical bottleneck, struggling with the scale and speed required for effective response \cite{b10}. The proliferation of visual data from cameras and satellites offers a solution, positioning deep learning as a necessary tool for automating this interpretation bottleneck \cite{b11, b12, b13, b14}.

However, applying standard deep learning models to this task is severely hampered by fundamental challenges. First, post-tornado imagery, filled with chaotic textures, debris, and fractured geometries, creates a significant "domain shift" from the clean, object-centric datasets on which models are typically pre-trained \cite{b15,b16}. Second, established engineering taxonomies like IN-CORE require models to solve competing multi-scale interpretation challenges: distinguishing fine-grained features (e.g., 10\% vs. 20\% roof damage) is a different task than identifying gross structural failure (e.g., a missing roof). Finally, real-world data is inherently afflicted with extreme class imbalance (e.g., a preponderance of undamaged structures) \cite{b17}.
Without a rigorous framework for systematic benchmarking and optimization, it remains unclear which models or optimization strategies are truly effective for this complex task. This research directly addresses these gaps by presenting a systematic experimental framework designed to deconstruct the problem of post-tornado damage recognition. Using our newly curated Quad-State Tornado Damage (QSTD) dataset as a testbed, we evaluate 79 publicly available deep learning architectures to answer two central questions: (1) Which architectural families are best suited for identifying tornado-specific damage patterns? (2) Which optimization strategies are most critical, and how do they interact with specific architectural families (Convolutional Neural Networks (CNNs) vs. Vision Transformers (ViT)) to impact performance?

Fundamentally, our work makes several contributions to the field:
\begin{enumerate}
\item	We introduce the QSTD dataset, a novel, large-scale, expertly labeled dataset based on the IN-CORE taxonomy, which serves as the foundation for this study.  
\item We establish a comprehensive computational framework for automating knowledge-intensive structural damage assessment, systematically evaluating how 79 deep learning architectures can learn to operationalize the component-based IN-CORE engineering taxonomy across 2,300+ controlled experiments, including an analysis of performance based on object-centric (ImageNet) versus scene-centric (Places365) pre-training.
\item	We conduct a large-scale optimization analysis revealing critical interactions between model architecture and training parameters. We demonstrate, for instance, that optimizer selection Stochastic Gradient Descent (SGD) vs. Adaptive Moment Estimation (Adam) can be more consequential than architectural choice \cite{b18}, providing a +25 to +38 point F1 uplift for the ViT and Swin Transformer families and fundamentally altering the performance hierarchy.
\item	We validate the robustness of our framework through a rigorous cross-event generalization test on the Tuscaloosa-Moore Tornado Damage (TMTD) dataset, demonstrating that our QSTD-trained model maintains significant predictive power across temporal and sensor-based domain shifts.
\end{enumerate}

The remainder of this paper is organized as follows: Section 2 reviews related work. Section 3 describes the datasets utilized. Section 4 details our methodological framework and architectures. Section 5 presents the experimental setup and evaluation metrics. Section 6 provides a systematic analysis of the experimental results. Section 7 discusses the implications of these findings and distills them into actionable deployment recommendations for automated tornado damage assessment systems and Section 8 concludes the paper with directions for future research.

\section{Related Works}
\subsection{Architectural Paradigms for Damage Assessment}
The application of deep learning to post-disaster damage assessment has rapidly evolved. Foundational CNNs, such as ResNet50, have been shown to be effective for fine-grained classification, achieving 90.28\% accuracy on post-tornado ground-level imagery for simplified, coarse-grained classification tasks \cite{b14}. A recent wave of research investigates Vision Transformers (ViTs) and Swin Transformers, which often outperform CNN baselines on satellite imagery \cite{b11,b19,b20} and may offer superior generalization to new disaster regions \cite{b21}.
This has spurred hybrid models, such as CTANet (fusing ConvNeXt and a Bilateral Hybrid Attention Transformer), which achieved a 95.74\% F1-score on building extraction surpassing CNNs like U-Net \cite{b22}, and other hybrids combining Inception-ResNet-v2 and EfficientNetV2B0 reaching 0.93 AUC \cite{b17}. Critically, most studies compare few architectures and focus on satellite-based segmentation. The effectiveness of these varied paradigms for fine-grained, ordinal damage classification from ground-level imagery remains a significant gap. Our work addresses this by providing the first known large-scale, systematic comparison of 79 open-source CNNs and Transformers on disaster imagery.
\subsection{Domain-Specific Challenges in Disaster Imagery}
Beyond architectural choice, model performance in disaster damage assessment is fundamentally limited by the unique characteristics of disaster data. Two most significant challenges are domain shift and class imbalance, both of which have driven extensive research in the field. In particular, domain shift arises when models trained on one region, structure type, or disaster event fail to generalize to new and unseen conditions. To combat this, recent work explores unsupervised domain adaptation (UDA). For example, Ahn et al. (2025) developed the DAVI framework, which leverages foundation models like SAM to adapt to new domains without labels, achieving a 0.68 F1-score on the 2023 Turkey earthquake data, significantly outperforming prior adaptation baselines \cite{b15}.
A key strategy to mitigate the initial domain gap is the choice of pre-training source. This typically involves a trade-off between object-centric datasets like ImageNet \cite{b23} and scene-centric datasets like Places365 \cite{b24}]. A 2020 benchmark by Ethan et al. directly compared these through an ablation study, finding that Places365 pre-training generally provided better mean average precision (mAP) for incident recognition tasks and was more suited to broad scene classification \cite{b16}. This highlights a complex relationship, which inspires our investigation into whether these scene-centric benefits transfer to the post-disaster assessment domain.
Finally, post-disaster data is notoriously long-tailed and imbalanced, with undamaged structures vastly outnumbering severe cases. To address this and related data scarcity, modern approaches are moving beyond simple re-weighting. For instance, Lagap \& Ghaffarian (2025) utilized Enhanced Super-Resolution GANs (ESRGAN) to generatively augment minority classes, demonstrating improved recall and raising ViT accuracy from 79\% to 84\% \cite{b25}. While these studies propose powerful solutions, their interaction with a wide array of architectures and optimizers is not well understood, motivating the systematic optimization in our framework.

\subsection{Optimization and Evaluation Strategies for Damage Assessment}
Adapting models to disaster imagery requires robust optimization beyond just selecting an architecture. Data augmentation, for example, is essential for imbalanced datasets. While classic geometric transforms provide modest gains, generative approaches like ESRGANs have improved F1-scores for minority classes by up to 18\% \cite{b26}. These improvements are driven by the tuning of hyperparameters (adjustable training settings such as learning rates and loss functions) that allow the training process to be tailored to specific objectives like overcoming class imbalance.
Optimizer selection is a primary consideration in this process.  While Adam is common for rapid convergence, SGD often provides superior performance. Hong et al. (2022) used a carefully tuned SGD configuration to achieve 77.49\% accuracy in earthquake damage classification \cite{b27}. Similarly, specialized loss functions are used to move beyond standard cross-entropy. Chen (2021) found that combining Focal Loss with a Dice loss significantly improved F1-scores (from 0.265 to 0.63) by forcing the model to focus on difficult, underrepresented damage states \cite{b28}. 
Perhaps the most significant methodological gap is in evaluation. Standard F1-score can be misleading as they treat all misclassifications with equal weight, which is ill-suited for the ordinal nature of damage scales. Recent research suggests that ordinal-aware metrics, such weighted kappa, provide a more reliable measure for assessing tasks with a natural progression of severity \cite{b29,b30}. This study addresses these gaps by systematically investigating hyperparameter interactions at scale and introducing ordinal metrics for a more meaningful assessment of model utility in disaster response.

\section{The Quad-State Tornado Damage (QSTD) Benchmark}
\subsection{Data Source and Taxonomy}
The QSTD dataset is derived from 4,800 geotagged, street-view images collected during the "Wave 1" field mission (three weeks post-event) following the historic December 2021 Quad-State tornado outbreak \cite{b10}. This ground-level perspective provides a realistic data source for field assessment but introduces challenges like occlusions and variable lighting.
Our classification framework extends the objective, component-based IN-CORE taxonomy (\Cref{tab:damage_taxonomy}), which defines four progressive damage states (DS1-DS4) based on observable architectural and structural failures \cite{b8}. To create a robust system for automated pipelines, we added two essential classes: DS0 (Undamaged) and DS5 (Debris/Non-Structural). The DS5 class is critical, enabling the model to identify non-assessable images (due to obstructions, debris, or lack of a structure), a necessary capability for real-world deployment.
\subsection{Curation and Final Statistics}
\begin{figure}[htbp]
    \centering
    \includegraphics[width=1\linewidth]{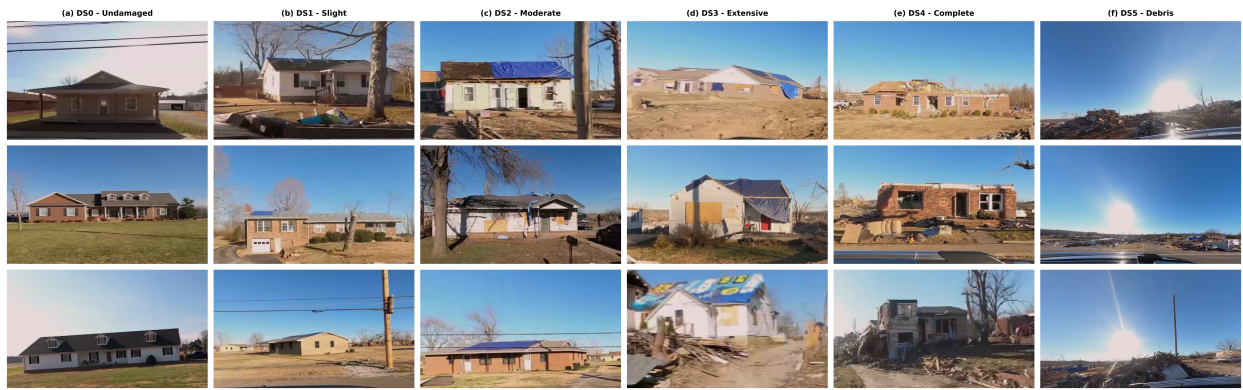}
    \caption{Example of QSTD images for the 6-class IN-CORE taxonomy}
    \label{fig:Figure_1}
\end{figure}

The benchmark was constructed via a rigorous two-phase protocol. First, two trained annotators independently cropped 5,583 building-centric images from the 4,800 wide-angle source frames and assigned initial damage labels. This increase in the total count resulted from the presence of multiple structures within single wide-angle views, requiring individual crops for each building. Annotators assigned initial damage labels to each of these instances. 
In Phase 2, the annotators conducted a full cross-validation and consensus review. This process involved verifying crop quality, removing duplicates, and resolving ambiguous boundary cases – most notably reclassifying 241 images from DS2 to DS3 to strictly align with IN-CORE criteria. A detailed summary of these curation adjustments and quality-control exclusions is provided in \Cref{app:annotation_protocol} and image samples are shown in \Cref{fig:Figure_1}.
The final QSTD benchmark comprises 5,517 curated images, partitioned via stratified sampling into training (70\%), validation (15\%), and test (15\%) sets. The dataset exhibits a realistic, long-tail class distribution (\Cref{fig:Figure_2}(a)), with DS0 (Undamaged) comprising 61.2\% of samples while the critical DS3 (Extensive) class represents only 6.7\%. This severe imbalance, along with variable image resolutions and the ground-level perspective, poses a significant and realistic challenge for model development.

\begin{table}[htbp]
\centering
\caption{IN-CORE Damage State Classification Criteria for Wood-Frame Residential Structures.}
\label{tab:damage_taxonomy}
\small
\begin{tabular}{@{}l>{\columncolor{green!15}}p{2.8cm}>{\columncolor{yellow!20}}p{2.8cm}>{\columncolor{orange!20}}p{2.8cm}>{\columncolor{red!20}}p{2.8cm}@{}}
\toprule
\textbf{Element} & \textbf{DS1 - Slight} & \textbf{DS2 - Moderate} & \textbf{DS3 - Extensive} & \textbf{DS4 - Complete} \\
\midrule
Roof Covering & 2--15\% of Roof Covering Damaged & 15--50\% of Roof Covering Damage & More than 50\% of Roof Covering Damaged & More than 50\% of Roof Covering Damaged (typically) \\
\midrule
& \multicolumn{4}{c}{\textit{AND/OR}} \\
\midrule
Window/Door & 1 window or door failure & 2 or 3 windows/doors failed & More than 3 windows/doors failed & More than 3 windows/doors failed (typically) \\
\midrule
& \multicolumn{4}{c}{\textit{AND/OR}} \\
\midrule
Roof Sheathing & No Roof Sheathing Failure & 1--3 sections of roof sheathing failed & More than 3 sections AND less than 35\% of roof sheathing failed & More than 35\% of roof sheathing failed \\
\midrule
& \multicolumn{4}{c}{\textit{AND/OR}} \\
\midrule
Roof-to-Wall Connection & No Roof-to-Wall Connection Failure & No Roof-to-Wall Connection Failure & No Roof-to-Wall Connection Failure & Roof-to-Wall Connection Failure \\
\bottomrule
\end{tabular}
\end{table}

\subsection{The Tuscaloosa-Moore Tornado Damage (TMTD) Test Set}
\begin{figure}[H]
    \centering
    \includegraphics[width=1\linewidth]{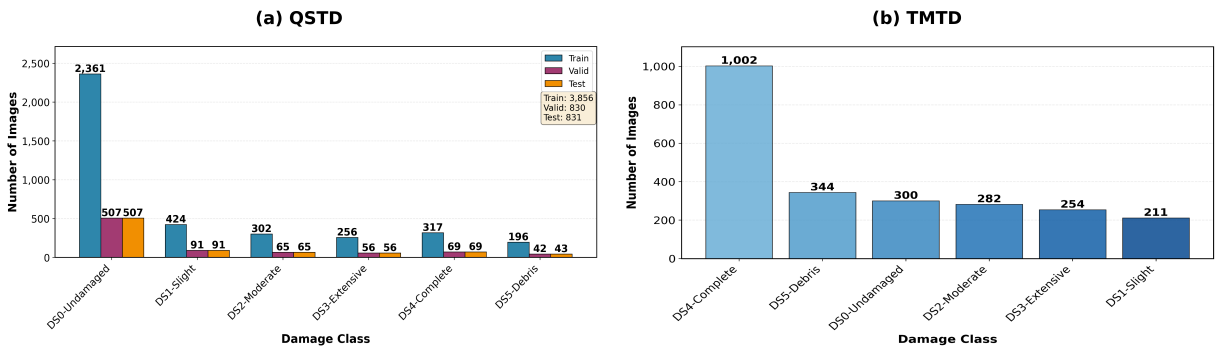}
    \caption{(a) Class distribution of the QSTD dataset across training, validation, and test splits, highlighting the pronounced long-tail imbalance. (b) Overall class distribution of the TMTD dataset used for cross-event evaluation}
    \label{fig:Figure_2}
\end{figure}

To validate cross-event generalization under realistic deployment conditions, we use the Tuscaloosa–Moore Tornado Damage (TMTD) dataset as a held-out zero-shot test set. TMTD contains 2,393 images from the April 27, 2011 Tuscaloosa EF4 tornado and the May 20, 2013 Moore EF5 tornado (\Cref{fig:Figure_2}(b) and \Cref{fig:Figure_3}) \cite{b31}. Importantly, no TMTD images are used during training, allowing a strict zero-shot assessment. Compared to QSTD, TMTD introduces substantial domain shifts: a temporal gap of nearly a decade, a camera modality shift (handheld digital cameras vs. vehicle-mounted 360° systems), and geographic variations affecting architectural styles. The final “champion” model selected from QSTD optimization is applied directly to TMTD without fine-tuning, mirroring operational conditions where a system must function immediately. Performance on TMTD therefore serves as the primary measure of whether the QSTD-optimized framework produces models that transfer reliably across time, location, and sensor conditions.

\section{Developed Methodology}
\subsection{Problem Formulation for Post-Tornado Damage Recognition}
The post-tornado damage recognition task is formulated as a supervised multi-class image classification problem. The objective is to learn a mapping function that accurately assigns one of $C$ predefined damage state classification labels to an input image. Let $\mathcal{X}$ denote the input space of all possible aerial or ground-level images depicting post-disaster scenes. Let $Y = \{y_1, y_2, \ldots, y_C\}$ be the set of $C = 6$ distinct damage classification labels, specifically: DS0 (Undamaged), DS1 (Slight), DS2 (Moderate), DS3 (Extensive), DS4 (Complete), and DS5 (Obstructed/Non-Structural). The fundamental goal is to learn a function $f: \mathcal{X} \rightarrow Y$ that maps an input image to its corresponding damage level as defined by the IN-CORE taxonomy introduced in Section 3.1.

Given a training dataset $\mathcal{D}_{\text{train}} = \{(x_i, y_i)\}_{i=1}^{N}$ consisting of $N$ image-label pairs, the learning task is to estimate the optimal parameters $\theta^*$ of a model $f_\theta$ that minimizes a loss function $\mathcal{L}$ over the training data:
\begin{equation}
\theta^* = \arg\min_{\theta} \frac{1}{N} \sum_{i=1}^{N} \mathcal{L}(f_\theta(x_i), y_i)
\label{eq:optimization}
\end{equation}
For multi-class classification, the standard loss function is the Categorical Cross-Entropy (CCE) loss. For a given sample ($x_i$  , $y_i$), the CCE loss is defined as:
\begin{equation}
\mathcal{L}_{\text{CCE}}(f_\theta(x_i), y_i) = -\sum_{c=1}^{C} y_{i,c} \log(p_{i,c})
\label{eq:cce_loss}
\end{equation}
Where $y_{i,c}$) is the one-hot ground-truth label (equals 1 if class is $c$, 0 otherwise). The term $p_{i,c} = P(y=c \mid x_i)$  represents the model’s predicted conditional probability that image $x_i$ belongs to class c. This is computed by applying the softmax function to the network's output logits $z_i$, which transforms the unbounded scores into a normalized probability distribution: 
\begin{equation}
p_{i,c} = \frac{e^{z_{i,c}}}{\sum_{j=1}^{C} e^{z_{i,j}}}
\label{eq:softmax}
\end{equation}

This transformation ensures that the predicted probabilities lie in the range [0,1] and sum to one, establishing the baseline training objective used to evaluate the model performance across all experiments.

\begin{figure}[H]
    \centering
    \includegraphics[width=1\linewidth]{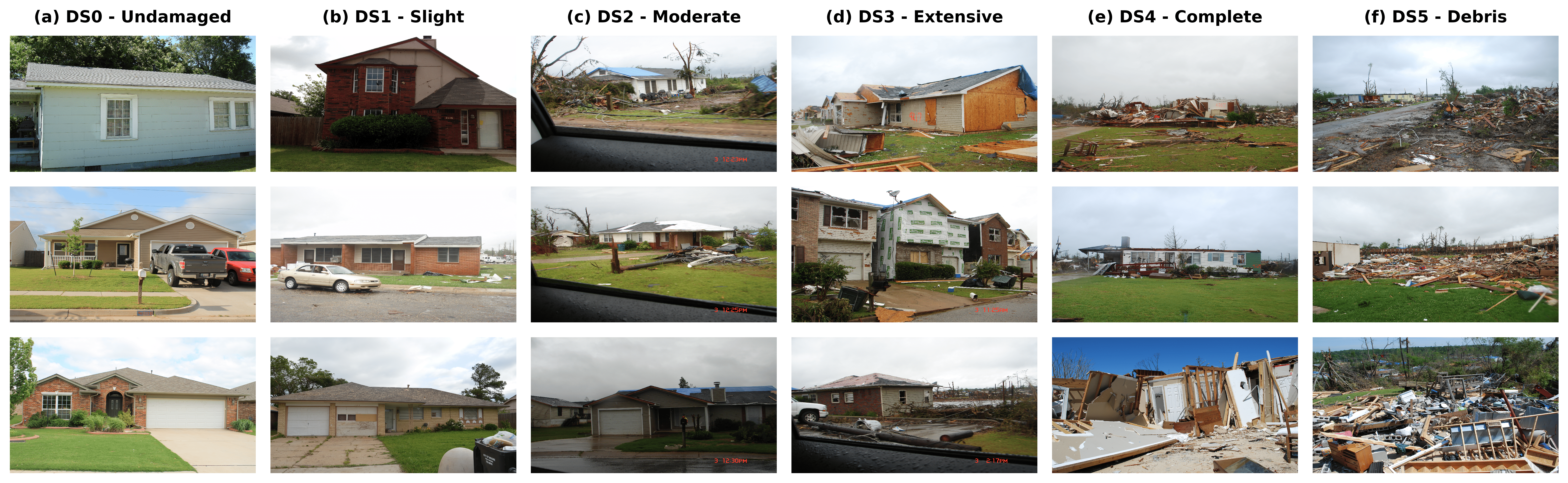}
    \caption{Example of TMTD images for the 6-class IN-CORE taxonomy}
    \label{fig:Figure_3}
\end{figure}

\subsection{Architectural Paradigms for Damage Recognition}\label{PM}
A core principle of this study is the exclusive use of publicly available, open-source architectures, ensuring transparency and reproducibility essential for high-stakes disaster response applications. Our framework evaluates 79 model variants spanning 16 architectural families (\Cref{app:model_list}), systematically connecting core design principles to the visual challenges of post-tornado damage recognition. We categorize these architectures into two primary paradigms, CNNs and Vision Transformers, enabling direct comparison of their fundamental approaches to this task.

\subsubsection{Convolutional Neural Network (CNNs)}
CNNs construct hierarchical image representations through learned local filters, making them inherently suited for identifying component-level damage features such as broken windows, missing shingles, or breaches in siding. We evaluate five keys CNN design philosophies:
\begin{itemize}
\item \textbf{Classic CNNs}: Foundational architectures like AlexNet \cite{b32} and the VGG family \cite{b33} established the principle that network depth is crucial for learning a rich feature hierarchy. Their stacked layers of simple 3x3 filters are effective at capturing basic damage textures and shapes.
\item \textbf{Residual Architectures}: To overcome the training degradation in very deep networks, architectures like ResNet \cite{b34}, ResNeXt \cite{b35}, and Wide ResNet \cite{b36} introduced identity shortcut connections. These enable the construction of much deeper models capable of learning the more complex and abstract patterns associated with severe structural damage. Similarly, DenseNet \cite{b37} promotes feature reuse through dense connectivity, which can be particularly effective for capturing the varied textures of debris and damaged materials.
\item	\textbf{Efficient Architectures}: For practical deployment in disaster response, computational efficiency is paramount. We evaluate a range of models designed for this purpose, including the EfficientNet family \cite{b38}, which systematically scales network dimensions for an optimal accuracy-to-computation trade-off. We also include lightweight models like MobileNet \cite{b39}, ShuffleNet \cite{b40}, and SqueezeNet \cite{b41}.
\item	\textbf{Neural Architecture Search (NAS) Designs}: This family includes models whose structures are optimized to capture features at multiple scales within a single block. GoogLeNet \cite{b42}, with its "Inception" module, was a key precursor. RegNet \cite{b43} uses a principled design space derived from NAS to create simple, effective network structures. We hypothesize that their ability to process information at multiple resolutions simultaneously is highly effective for damage assessment, where both fine-grained (e.g., individual missing shingles) and coarse features (e.g., total roof collapse) coexist.
\item	\textbf{Modernized CNNs}: Representing the current state-of-the-art in pure convolutional design, ConvNeXt \cite{b44} modernizes ResNet by incorporating principles from Vision Transformers. It serves as the ultimate CNN baseline, testing the performance limits of the convolutional paradigm on our task.
\end{itemize}

\subsubsection{Vision Transformers (ViTs)}
Vision Transformers employ self-attention mechanisms to model relationships between all image regions, capturing global scene context rather than local patterns. This global contextual reasoning is particularly advantageous when damage is widespread (e.g., complete roof loss in DS4) or when distinguishing target structures from chaotic debris fields requires holistic scene understanding.
The ViT paradigm divides images into fixed-size patches, which are linearly embedded with positional encodings and processed through Multi-Head Self-Attention layers. Each patch representation is updated via weighted aggregation across all patches:

\begin{equation}
\text{Attention}(Q, K, V) = \text{softmax}\left(\frac{QK^T}{\sqrt{d_k}}\right)V
\label{eq:attention}
\end{equation}

where $Q$ (queries), $K$ (keys), and $V$ (values) are learned linear projections of the input patch embeddings. The dot product $QK^T$ computes the pairwise similarity between all query--key pairs, while $\sqrt{d_k}$ acts as a scaling factor based on the key dimension $d_k$ to prevent vanishing gradient in the softmax function. Finally, the $\text{softmax}(\cdot)$ operation normalizes these similarity scores into attention weights, determining how strongly each patch contributes to the updated representation. This formulation enables the model to learn long-range spatial dependencies across the entire image.

We evaluate two ViT families in this study:
\begin{itemize}
    \item	\textbf{Global-Attention Transformers}: The original Vision Transformer (ViT) \cite{b45} applies self-attention globally across all patches, providing maximum contextual awareness at high computational cost, making it particularly suited for assessing overall structural integrity and distinguishing damage severity patterns.
    \item	\textbf{Hierarchical Transformers}: To improve efficiency and build a multi-scale understanding akin to CNNs, models like the Swin Transformer \cite{b46} and MaxViT \cite{b47} compute self-attention within local windows. By shifting these windows and merging patch representations across layers, they achieve linear complexity while still capturing powerful contextual relationships, offering a compelling hybrid of both paradigms.
\end{itemize}

The complete specifications for all 79 model variants are provided in \Cref{app:model_list} of the Supplementary Material. 

\subsection{Experimental Framework and Optimization Strategy }
\label{sec:others}
To address the research questions posed, we design a multi-stage experimental framework to systematically deconstruct the factors influencing model performance on post-tornado damage recognition. 
\subsubsection{Problem Formulation}\label{CPOM}
Let $\mathcal{A}$ denote the set of publicly available models under evaluation and let $A \in \mathcal{A}$ represent a specific architecture. For each architecture $A$, we seek to find the optimal learnable parameters $\theta^*$, and hyperparameter configuration $\psi^*$ from the search space $\Psi$ that maximizes a performance metric $\mathcal{P}$ (specifically, macro F1-score) on the validation set, $\mathcal{D}_{\text{val}}$:

\begin{equation}
(A^*, \psi^*) = \arg\max_{A \in \mathcal{A}, \psi \in \Psi} \mathcal{P}(f_{\theta^*(A,\psi)}, \mathcal{D}_{\text{val}})
\label{eq:global_optimization}
\end{equation}

where $A^*$ and $\psi^*$ represent the optimal architecture and hyperparameter configuration, respectively.

The term $\theta^*(A,\psi)$ represents the optimal model parameters (weights and biases) for a specific architecture $A$ and configuration $\psi$, obtained by minimizing the training loss $\mathcal{L}$ (\Cref{eq:cce_loss}):

\begin{equation}
\theta^*(A,\psi) = \arg\min_{\theta} \frac{1}{|\mathcal{D}_{\text{train}}|} \sum_{(x,y) \in \mathcal{D}_{\text{train}}} \mathcal{L}(f_A(x;\theta), y; \psi)
\label{eq:inner_optimization}
\end{equation}

where $f_A(x;\theta)$ denotes the output of architecture $A$ with parameters $\theta$ for input $x$, and the loss function $\mathcal{L}$ may depend on hyperparameters in $\psi$.

\subsubsection{Multi-Stage Experimental Design}\label{MFIK}
\begin{figure}[htpb]
    \centering
    \includegraphics[width=1\linewidth]{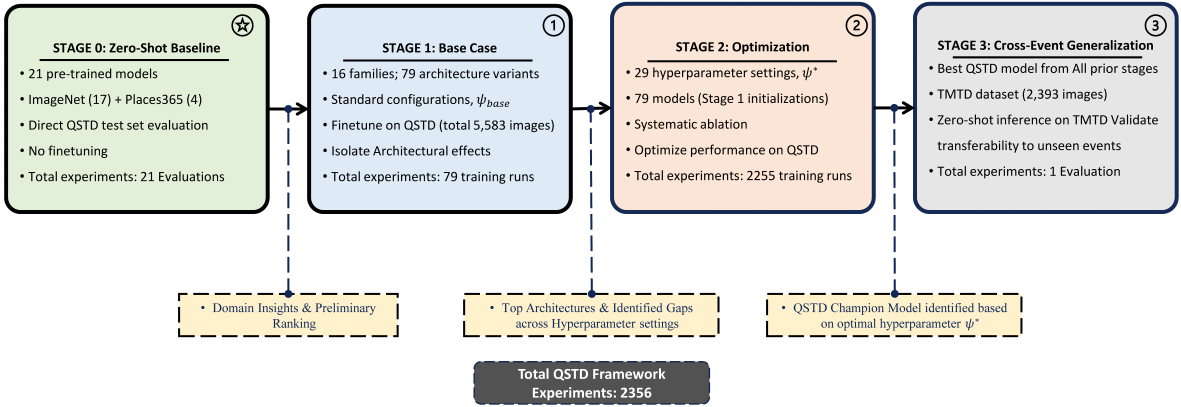}
    \caption{The four-stage experimental design, from zero-shot baseline experiment to cross-event generalisation evaluation.}
    \label{fig:Figure_4}
\end{figure}

Given the vastness of the joint search space ($A$,$\psi$), we employ a structured four-stage approach to make the optimization problem tractable while isolating the effects of architectural design from hyperparameter tuning. This staged methodology, illustrated in \Cref{fig:Figure_4}, enables systematic ablation studies, and provides clear insights into the relative importance of different design decisions.

\paragraph{Stage 0: Zero-Shot Baseline.} In the initial stage, we establish a domain-shift baseline by evaluating 21 models (17 ImageNet-1K and 4 Places365 pre-trained variants) directly on the QSTD test set without any fine-tuning. This stage quantifies the domain gap between general-purpose benchmarks and post-disaster assessment and provides a comparative evaluation of transfer learning from object-centric (ImageNet) versus scene-centric (Places365) sources. [Total experiments: 21 evaluations]

\paragraph{Stage 1: Base Case Architectural Comparison.} This stage addresses our primary research question regarding architectural suitability. We evaluate all 79 model variants (\Cref{app:model_list}), initialized with pre-trained weights, by fine-tuning them with a standard base configuration, $\psi_{base}$. This controlled experiment isolates the impact of architectural inductive biases (e.g., local CNN fields vs. global ViT attention) from the effects of hyperparameter tuning. [Total experiments: 79 training runs]

\paragraph{Stage 2: Systematic Hyperparameter Optimization.} Building upon Stage 1 insights, this stage investigates the impact of 29 distinct hyperparameter configurations (detailed in \Cref{tab:model_variants} and \Cref{tab:hyperparameters}) across all 79 architectures. This ablation approach, resulting in 2,291 training runs, isolates the contribution of each hyperparameter (e.g., loss function, optimizer, input resolution) and quantifies its impact across diverse architectural families. [Total experiments: 2,291 training runs]

\paragraph{Stage 3: Cross-Event Generalization Evaluation.} To validate real-world robustness, we employ the TMTD dataset as a held-out test set. TMTD comprises 2,393 images from the 2011 Tuscaloosa and 2013 Moore tornadoes, introducing critical temporal, sensor, and geographic domain shifts. The champion model from our QSTD optimization is evaluated on TMTD in a zero-shot setting, mirroring a realistic deployment scenario. [Total experiments: 1 evaluation]
This systematic exploration ensures that our performance findings are rooted in optimized configurations rather than default settings. The comprehensive search space, including the base case configurations and the domain-specific rationale for each optimization factor, is summarized in \Cref{tab:hyperparameters}.

\section{Experimental Setup and Evaluation}

\paragraph{Implementation Details.} All experiments were conducted using PyTorch 2.6.0 on NVIDIA A100 GPUs (40GB/80GB VRAM). Models were initialized with pre-trained weights from two sources: ImageNet-1K (object-centric recognition) for 75 architectures and Places365 (scene-level understanding) for 4 architectures (AlexNet, DenseNet161, ResNet18, ResNet50). This dual-source comparison enables assessment of whether scene-level features transfer more effectively to disaster imagery than object-centric representations. For ImageNet models, we used IMAGENET1K\_V1 weights for Stage 1 baseline comparisons and IMAGENET1K\_V2 weights (where available) for Stage 2 hyperparameter optimization, as the V2 recipe incorporates improved training protocols that enhance generalisation (Wightman et al., 2021). Training proceeded for a maximum of 100 epochs using the Adam optimizer (lr=0.001, $\beta_1=0.9$, $\beta_2=0.999$) with early stopping triggered after 10 epochs without improvement in validation macro F1-score. 

\paragraph{Classification Performance Metrics.} QSTD exhibits severe class imbalance (\Cref{fig:Figure_3}), making overall accuracy an unreliable indicator of model quality. Accordingly, we adopt macro-averaged F1-score as our primary metric. The F1-score, defined as the harmonic mean of precision (positive predictive value) and recall (sensitivity), provides a balanced measure of classification performance. Importantly, the macro-average treats all damage classes equally, ensuring that error on minority classes (DS3: 6.7\%, DS4: 8.2\%) are not overshadowed by strong performance on the majority class. To further characterize model behavior, we report per-class precision and recall to identify specific error modes. For practical deployment considerations, we additionally evaluate computational efficiency using parameter count (M), floating-point operations (GFLOPs), and single-image inference time (ms) measured on A100 hardware.
\paragraph{Domain-Specific Metric.} Recognizing the ordinal nature of the IN-CORE damage taxonomy, we introduce Ordinal Top-k Accuracy, a metric specifically adapted for evaluating classifiers where near-miss predictions are more acceptable than distant errors. Under this metric, a prediction is considered correct if any of the model's top-k predicted classes fall within one ordinal level of the ground truth:

\begin{equation}
\text{Acc}_{\text{ord}}^{(k)} = \frac{1}{N_{\text{eff}}} \sum_{i=1}^{N} \mathds{1}\left(d_i \leq 1 \land y_i \in \{0, 1, 2, 3, 4\}\right)
\label{eq:ordinal_accuracy}
\end{equation}

In \Cref{eq:ordinal_accuracy} above, $d_i = \min_{j \in \text{Top-}k(f(x_i))} |y_i - j|$ denotes the minimum ordinal distance between the true label $y_i$ and any of the Top-$k$ predicted class indices $j$ for sample $x_i$. The indicator function $\mathds{1}(\cdot)$ returns 1 if the condition is met, $N_{\text{eff}}$ is the number of samples whose true label $y_i$ belongs to the ordinal classes (DS0--DS4). $\text{Top-}k(f(x_i))$ represents the set of indices for the $k$ highest probability predictions. For this study, we primarily focus on Ordinal Top-1 Accuracy ($k=1$).

This formulation explicitly quantifies a model's ability to identify the correct severity level within one tolerance level, acknowledging that misclassifying DS2 as DS3 is far less consequential than misclassifying DS0 as DS4. We exclude the nominal DS5 class from ordinal calculations. Although several ordinal metrics exist in broader literature, this specific Top-k adaptation for damage assessment offers a nuanced and domain-aligned evaluation. Complementary ordinal confusion matrices further illustrate model behavior by visually separating tolerable adjacent-class errors from critical multi-state misclassifications.

\begin{table}[htbp]
\centering
\caption{Evaluated Architectural Families and Model Variants Across Experimental Stages.}
\label{tab:model_variants}
\small
\renewcommand{\arraystretch}{1.5}  % 1.5x normal height
\setlength{\tabcolsep}{5pt}  % default is 6pt
\begin{tabular}{@{}lcccccc@{}}
\toprule
\textbf{Architectural} & \textbf{Pre-training} & \textbf{Representative Models} & \textbf{Stage 0} & \textbf{Stage 1} & \textbf{Stage 2} & \textbf{Stage 3} \\
\textbf{Family} & \textbf{Sources} & \textbf{(Stage 0)} & \textbf{(Zero-Shot)} & \textbf{(Base Case)} & \textbf{(Optimization)\textsuperscript{§}} & \textbf{(TTD)} \\
\midrule
AlexNet & \dag, $\bigstar$ & AlexNet~\cite{b32} & 2 & 2 & 58 & \xmark \\
ConvNeXt & \dag & ConvNext-Small~\cite{b44} & 1 & 4 & 116 & \cmark \\
DenseNet & \dag, $\bigstar$ & DenseNet161~\cite{b37} & 2 & 5 & 145 & \xmark \\
EfficientNet & \dag & EfficientNet-B3~\cite{b38} & 1 & 12 & 348 & \xmark \\
GoogLeNet & \dag & GoogLeNet~\cite{b42} & 1 & 1 & 29 & \xmark \\
MaxViT & \dag & MaxViT-t~\cite{b47} & 1 & 1 & 23\textsuperscript{‡} & \xmark \\
MobileNet & \dag & MobileNet-v2~\cite{b39} & 1 & 4 & 116 & \xmark \\
RegNet & \dag & RegNet-x-16gf~\cite{b43} & 1 & 10 & 290 & \xmark \\
ResNet & \dag, $\bigstar$ & ResNet18, ResNet50~\cite{b34} & 4 & 10 & 290 & \xmark \\
ResNeXt & \dag & ResNeXt50-32x4d~\cite{b35} & 1 & 5 & 145 & \xmark \\
ShuffleNet & \dag & ShuffleNet-v2-x1.5~\cite{b40} & 1 & 4 & 116 & \xmark \\
SqueezeNet & \dag & SqueezeNet1.0~\cite{b41} & 1 & 1 & 29 & \xmark \\
Swin Transformer & \dag & Swin-B~\cite{b46} & 1 & 6 & 174 & \xmark \\
VGG & \dag & VGG16~\cite{b33} & 1 & 8 & 232 & \xmark \\
Vision Transformer & \dag & ViT-L-16~\cite{b45} & 1 & 5 & 115\textsuperscript{‡} & \xmark \\
Wide ResNet & \dag & Wide-ResNet50-2~\cite{b36} & 1 & 1 & 29 & \xmark \\
\midrule
\textbf{Total Experiments (2356)} & & & 21 & 79 & 2,255 & 1 \\
\bottomrule
\multicolumn{7}{l}{\scriptsize \textsuperscript{§} Total Hyperparameter Variations (excluding base case): 29} {\scriptsize \textsuperscript{‡} Excludes input resolution experiment due to patch compatibility and computational constraints} \\
\multicolumn{7}{l}{\scriptsize \dag~ImageNet \quad \quad \quad \quad \quad  $\bigstar$~Places365 \quad \quad \quad \quad \cmark~`Champion' model from stage 1 and 2 optimizations}
\end{tabular}
\end{table}

\section{Results and Analysis}
\subsection{Stage 0: Zero-Shot Transfer Learning (Quantifying the Domain Gap)}
To quantify the domain shift between general-purpose image classification and tornado damage assessment, we evaluated 21 pre-trained convolutional and transformer architectures (17 ImageNet-1K and 4 Places365) directly on the QSTD test set without fine-tuning. The zero-shots results, shown in \Cref{tab:zero_shot}, reveal an extensive domain gap that severely limits the transferability of conventional visual representations to post-disaster imagery.

\begin{table}[H]
\centering
\caption{Comprehensive Hyperparameter Search Space and Optimization Strategy.}
\label{tab:hyperparameters}
\small
\renewcommand{\arraystretch}{1.4}
\setlength{\tabcolsep}{3.5pt}
\begin{tabular}{@{}lp{2.2cm}p{3.5cm}cp{5.5cm}@{}}
\toprule
\textbf{Hyperparameter} & \textbf{Base Case} & \textbf{Search Space} & \textbf{Variations\textsuperscript{‡}} & \textbf{Domain-Specific Goal} \\
& $(\psi_{\text{base}})$ & $(\Psi)$ & & \\
\midrule
Activation function & Model Default* & \{ReLU, LeakyReLU, GELU, SiLU, ELU\} & 5 & Investigate impact of non-linearity on learning complex damage textures and preventing dying neurons \\
% \midrule
Batch Size & 64 & \{64, 32, 16, 8, 4\} & 4 & Balance gradient stability with noise-induced regularization; assess impact on rare class representation \\
% \midrule
Optimizer & Adam & \{Adam, SGD (momentum=0.9)\} & 1 & Compare adaptive vs. momentum-based optimization for disaster imagery \\
% \midrule
Regularization & None & \{None, Dropout with $p \in \{0.2, 0.3, 0.5\}$\} & 3 & Combat overfitting to dataset-specific debris patterns and lighting artifacts \\
% \midrule
Input resolution & $224 \times 224$ & \{$160^2$, $192^2$, $224^2$, $256^2$, $384^2$, $448^2$, $512^2$\}\textsuperscript{†} & 6 & Enable multi-scale feature recognition from fine-grained (windows) to gross (structural) damage \\
% \midrule
Data augmentation & Basic & \{None, Basic, Standard, Advanced, Heavy\} & 4 & Build robustness to vehicle-mounted camera variations (angle, lighting, occlusion) \\
% \midrule
LR Scheduler & None & \{None, Step Decay, Cosine Annealing\} & 2 & Prevent premature convergence and enable fine-grained late-stage optimization \\
% \midrule
Learning rate & $1 \times 10^{-3}$ & \{$1 \times 10^{-3}$, $3 \times 10^{-3}$, $5 \times 10^{-3}$, $1 \times 10^{-4}$\} & 3 & Tune optimization step size for diverse architectural families \\
% \midrule
Loss function & CCE & \{CCE, Focal Loss ($\gamma = 5$, $\alpha = 0.5$)\} & 1 & Address severe class imbalance (DS0: 61\% vs. DS3: 7\%, DS4: 8\%) \\
\bottomrule
\multicolumn{5}{l}{\scriptsize \textsuperscript{‡} 29 variations excluding base case value, evaluated in Stage 2 \quad \quad \scriptsize * Default activations vary by family: ReLU (ResNet, VGG, AlexNet), ReLU6 (MobileNetV2),} \\
\multicolumn{5}{l}{\scriptsize Hardswish (MobileNetV3), SiLU (EfficientNet), GELU (ViT, Swin, ConvNeXt) \quad \scriptsize \textsuperscript{†} Batch size reduced to 16 for EfficientNet-B7 (Res$\geq$256), EfficientNet-V2-L} \\
\multicolumn{5}{l}{\scriptsize (Res$\geq$384), EfficientNet-V2-M (Res=512) and RegNet\_Y\_128GF (Res$\geq$192) due to GPU memory constraints (40/80GB VRAM limit)}
\end{tabular}
\end{table}

\begin{figure}[htbp]
    \centering
    \includegraphics[width=1\linewidth]{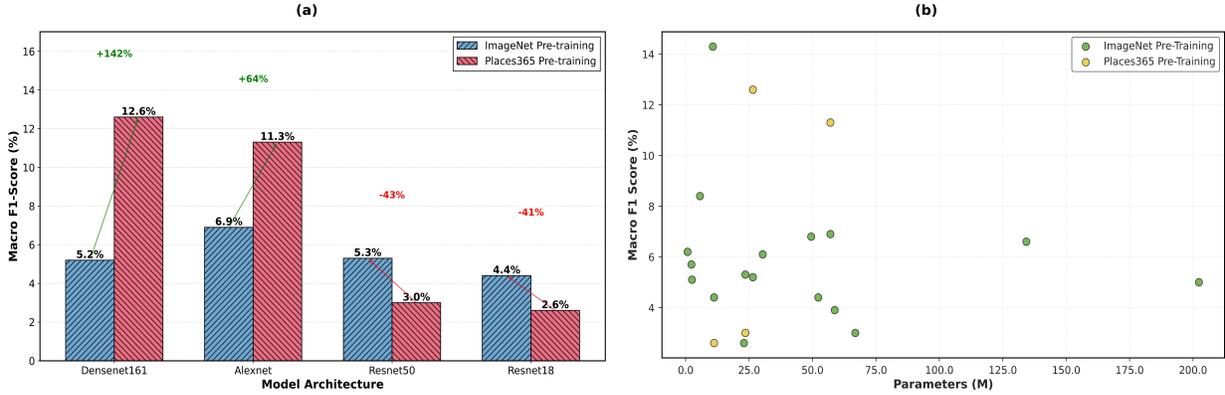}
    \caption{(a) Zero-Shot Transfer Performance by Pre-training Source. (b) Scatter Plot of Macro F1 vs. Parameters for all 21 zero-shot models showing no correlation between parameter count and F1.}
    \label{fig:Figure_5}
\end{figure}

\begin{figure}
    \centering
    \includegraphics[width=1\linewidth]{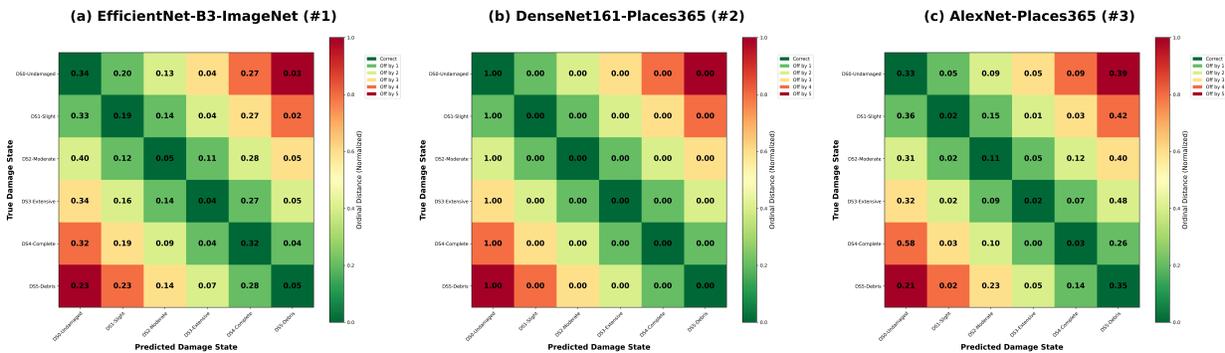}
    \caption{Ordinal confusion matrices for the top three zero-shot models on the QSTD test set. Color indicates ordinal distance (Dark Green = correct, Green = $\pm 1$, Light Green = $\pm2$, Yellow = $\pm 3$, Orange = $\pm 4$, Red = $\pm 5$). (a) EfficientNet-B3 (ImageNet) shows scattered predictions. (b) AlexNet (Places365) overpredicts DS0 and DS5. (c) DenseNet161 (Places365) collapses to DS0-Undamaged}
    \label{fig:Figure_6}
\end{figure}

\paragraph{Insight 1: Massive Domain Gap Confirmed.} Zero-shot transfer resulted in near-random performance across most architectures with 19 out of 21 models achieving a macro F1-score below 10\%. As summarized in \Cref{tab:zero_shot}, the average macro F1-score was only 6.0\% for ImageNet models and 9.7\% for Places365 models, compared with 75–85\% ImageNet validation accuracies reported for the same architectures. This 69+ percentage point drop quantifies the fundamental mismatch between natural object recognition and structural damage assessment. Even state-of-the-art architectures such as ViT-L/16 (5.0\% F1), Swin-B (3.9\% F1), and ResNet-50 (5.3\% F1) failed to generalize, highlighting that pre-training on object-centric datasets does not confer robustness to disaster imagery's unique visual characteristics. The best zero-shot performance, achieved by EfficientNet-B3 (14.3\% F1), still falls far below operational thresholds, confirming that fine-tuning is essential for domain adaptation.
\paragraph{Insight 2: Architectural Design Dictates the Optimal Pre-Training Source.} Comparing pre-training sources reveals a beneficial, though not universal, trend that depends strongly on each model’s architectural design. As depicted in \Cref{fig:Figure_5}(a) showing the zero-shot performance by pretraining source, DenseNet161 and AlexNet benefited substantially from the scene-centric features of Places365, improving by 142\% and 64\% over their ImageNet counterparts. Their dense connectivity and broad receptive fields appear well suited for leveraging contextual, scene-level cues. In contrast, the ResNet family favored the object-centric features of ImageNet, with both ResNet-50 and ResNet-18 outperforming their Places365 versions, suggesting that residual architectures transfer object-level knowledge more effectively. This architectural split indicates that the optimal pre-training source depends on a model’s inductive biases. The top performer, EfficientNet-B3 (14.3\% F1, ImageNet-trained), further reinforces this point. Its compound scaling design yields highly transferable representations and demonstrates that architectural efficiency can outweigh pre-training source alignment.
\paragraph{Insight 3: Latent Ordinal Understanding Despite Poor Classification.} Although macro F1 scores were uniformly low, several models demonstrated surprisingly high ordinal accuracy. ResNet-50-ImageNet, for instance, achieved only 5.3\% macro F1 but 83.9\% ordinal Top-1 accuracy. Similar patterns were observed for DenseNet161-Places365 (12.6\% F1, 75.9\% ordinal) and MaxViT-T (6.1\% F1, 64.3\% ordinal). This quantitative gap indicates that although models fail at exact classification, their incorrect predictions often fall near the true damage state rather than being randomly distributed across all classes. However, visual inspection of the confusion matrices (\Cref{fig:Figure_6}) reveals complex and varied failure patterns, including mode collapse in DenseNet161 and scattered misclassifications in EfficientNet-B3, instead of a clean diagonal concentration of adjacent errors. These observations suggest that while the feature space encodes some ordinal structure, the zero-shot classification heads cannot map these features reliably, resulting in visually chaotic confusion patterns. This reinforces the need for fine-tuning to properly calibrate the classification layer to the learned representations.
\paragraph{Insight 4: Model Scale Provides No Advantage Under Domain Shift.} Parameter count showed no correlation with zero-shot performance across the 21 models, as visualized in \Cref{fig:Figure_5}(b). Lightweight models such as SqueezeNet1.0 (0.7M parameters, 6.2\% F1) performed comparably to large transformers like ViT-L/16 (202.3M parameters, 5.0\% F1). Interestingly, EfficientNet-B3 achieved the highest F1 score with only 10.7M parameters, 12.6 times fewer than VGG16 (134.3M) and 18.9 times fewer than ViT-L/16 (202.3M). The absence of scaling benefit highlights that model capacity cannot compensate for the fundamental feature mismatch between pre-training and target domains. 
The zero-shot analysis provides clear directives for our subsequent experimental stages. First, the pronounced domain gap confirms that domain-specific fine-tuning (Stage 1) is essential for achieving meaningful performance. Second, the complex interaction between architecture and pre-training source motivates a broad evaluation across diverse architectural families, as the optimal combination is not predictable a priori. Third, the presence of some latent ordinal knowledge validates our use of ordinal metrics and suggests that fine-tuning should focus on adapting the classification head to leverage the already transferable backbone features. Finally, the absence of correlation between model scale and zero-shot success underscores the importance of systematic hyperparameter optimization (Stage 2), as performance gains are more likely to result from refined configurations than from brute-force scaling. These insights collectively shape the investigation in the following stages.

\begin{table}[H]
\centering
\caption{Zero-Shot Transfer Performance of Representative Models on QSTD Test Set.}
\label{tab:zero_shot}
\small
\renewcommand{\arraystretch}{1.08}
\begin{tabular}{@{}clcccccc@{}}
\toprule
\textbf{Rank\textsuperscript{‡}} & \textbf{Model} & \textbf{Pre-training} & \textbf{Macro-F1} & \textbf{Ordinal Top-1} & \textbf{Standard\textsuperscript{¶} Top-1} & \textbf{Param\textsuperscript{‖}} & \textbf{Inference} \\
& & & \textbf{(\%)} & \textbf{Acc. (\%)} & \textbf{Acc. (\%)} & \textbf{(M)} & \textbf{Time (ms)} \\
\midrule
1 & EfficientNet-B3 & \dag & \textbf{14.3} & 50.5 & 26.1 & 10.7 & 13.3 \\
2 & DenseNet161 & $\bigstar$ & 12.6 & 75.9 & \textbf{61.0} & 26.5 & 21.5 \\
3 & AlexNet & $\bigstar$ & 11.3 & 33.5 & 23.2 & 57.0 & \textbf{0.6} \\
4 & GoogLeNet & \dag & 8.4 & 14.8 & 9.4 & 5.6 & 7.1 \\
5 & AlexNet & \dag & 6.9 & 39.5 & 8.3 & 57.0 & \textbf{0.6} \\
6 & ConvNeXt-small & \dag & 6.8 & 28.6 & 8.1 & 49.4 & 9.6 \\
7 & VGG16 & \dag & 6.6 & 8.0 & 7.2 & 134.3 & 1.5 \\
8 & SqueezeNet1.0 & \dag & 6.2 & 49.9 & 10.2 & 0.7 & 1.7 \\
9 & MaxViT-T & \dag & 6.1 & 64.3 & 11.2 & 30.4 & 23.7 \\
10 & MobileNetV2 & \dag & 5.7 & 28.9 & 7.9 & 2.2 & 5.4 \\
11 & ResNet50 & \dag & 5.3 & \textbf{83.9} & 14.4 & 23.5 & 6.2 \\
12 & DenseNet161 & \dag & 5.2 & 45.6 & 9.0 & 26.5 & 21.6 \\
13 & ShuffleNet-V2-x1.5 & \dag & 5.1 & 26.3 & 8.2 & 2.5 & 6.7 \\
14 & ViT-L-16 & \dag & 5.0 & 24.4 & 6.4 & \textbf{202.3} & 14.1 \\
15 & ResNet18 & \dag & 4.4 & 6.1 & 6.1 & 11.2 & 2.5 \\
16 & RegNet-X-16GF & \dag & 4.4 & 33.0 & 8.3 & 52.2 & 8.6 \\
17 & Swin-B & \dag & 3.9 & 28.0 & 7.8 & 58.7 & 20.5 \\
18 & ResNet50 & $\bigstar$ & 3.0 & 9.4 & 5.5 & 23.5 & 6.2 \\
19 & Wide-ResNet50-2 & \dag & 3.0 & 16.1 & 8.2 & 66.8 & 6.3 \\
20 & ResNet18 & $\bigstar$ & 2.6 & 24.6 & 7.6 & 11.2 & 2.5 \\
21 & ResNeXt50-32x4d & \dag & 2.6 & 15.9 & 8.3 & 23.0 & 6.1 \\
\midrule
& ImageNet Average & \dag & 5.9 & 33.2 & 9.7 &  & 9.2 \\
& Places365 Average & $\bigstar$ & \textbf{7.4} & \textbf{35.8} & \textbf{24.3} &  & \textbf{7.7} \\
\bottomrule
\multicolumn{8}{l}{\scriptsize \textsuperscript{‡} Models ranked by macro F1-score \quad \quad \quad \quad \quad \quad  \scriptsize \dag~ImageNet-1K \quad \quad \quad \quad \quad \quad \quad  $\bigstar$~Places365 \quad \quad \quad \quad \quad \quad  \scriptsize \textsuperscript{¶} Standard Top-1 Accuracy (exact match only)} \\
\multicolumn{8}{l}{\scriptsize \textsuperscript{‖} Parameter counts include the model backbone and final classification layer adapted for 6 classes} \\
\end{tabular}
\end{table}


\subsection{Stage 1: Fine-Tuned Architectural Base Case Performance}
\begin{figure}
    \centering
    \includegraphics[width=1\linewidth]{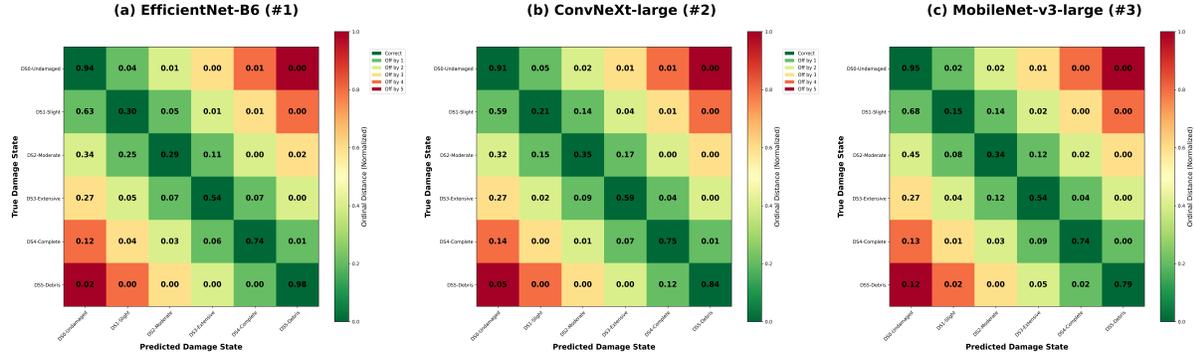}
    \caption{Ordinal confusion matrices of top-performing Stage 1 models for top three architecture families on the QSTD test set}
    \label{fig:Figure_8}
\end{figure}

\begin{figure}
    \centering
    \includegraphics[width=0.82\linewidth]{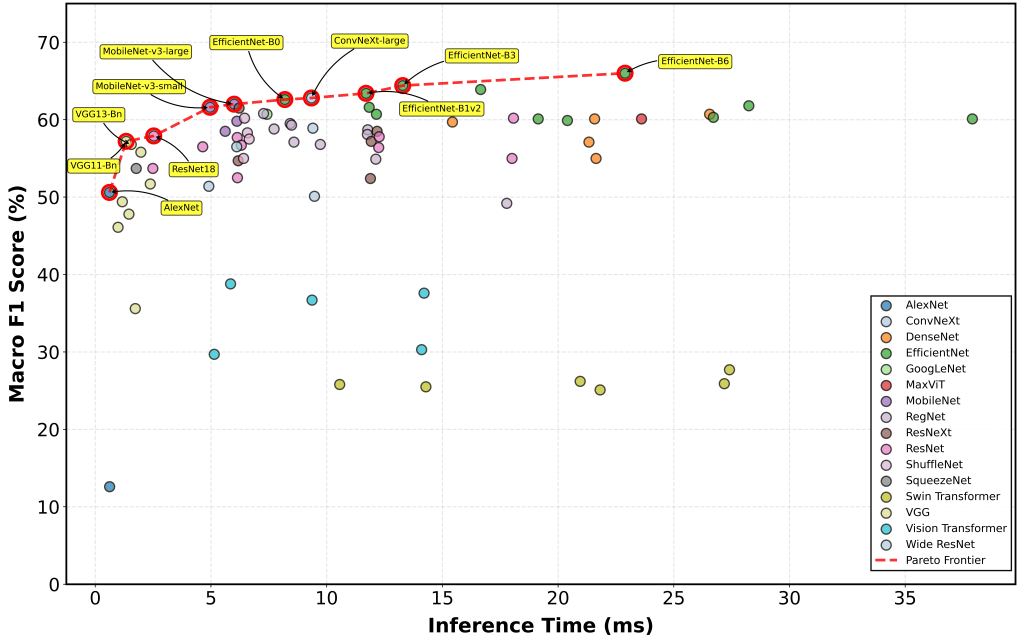}
    \caption{Scatter plot of Macro F1 vs. Inference Time for all 79 fine-tuned models showing the Pareto frontier}
    \label{fig:Figure_9}
\end{figure}

Moving beyond the zero-shot analysis, Stage 1 evaluates the performance of all 79 model variants on the QSTD training set using the base-case configuration ($\psi_{base}$). This stage establishes a comprehensive architectural baseline, quantifying how fine-tuning mitigates the severe domain gap and revealing systematic performance differences across model families.
\paragraph{Substantial Performance Gains through Fine-Tuning.} Fine-tuning produced a dramatic performance increase across all metrics, effectively bridging the zero-shot domain gap. Whereas zero-shot models averaged below 10\% Macro F1, fine-tuned architectures achieved an average of approximately 54\%, confirming that domain-specific adaptation is indispensable for post-disaster imagery analysis. The top-performing model, EfficientNet-B6 (ImageNet), reached 66.0\% Macro F1, 91.5\% Ordinal Top-1 Accuracy, and 77.7\% Standard Top-1 Accuracy, establishing a strong operational baseline for the QSTD task. \Cref{fig:Figure_8} shows the confusion matrices of the top three performing architectural families from stage 1. 
\paragraph{CNNs Show Superior Baseline Performance.} A clear hierarchy emerges when comparing architectural paradigms in the base case fine-tuning: Convolutional Neural Networks (CNNs) significantly outperform Vision Transformers (ViTs). As shown in \Cref{fig:Figure_7}(a), the best fine-tuned ViT (ViT-B-32, 38.8\% F1) and Swin Transformer (Swin-V2-S, 27.7\% F1) variants achieve Macro F1 scores markedly lower than top-performing CNNs, which exceed 60\% F1 under identical training conditions. This performance gap indicates that the inductive biases inherent in CNNs (spatial locality and translation equivariance) provide a more effective foundation for learning ground-level damage features than the global attention mechanisms of transformers when using a standard fine-tuning protocol. 
\paragraph{EfficientNet Leads, Lightweight Models Excel in Efficiency.} Analyzing the top performers reveals a striking dominance by the EfficientNet family. Variants of EfficientNet occupy six of the top ten positions in the baseline fine-tuned rankings (\Cref{app:stage1_results} ), including the overall top performer, EfficientNet-B6 (66.0\% Macro F1). This consistent trend strongly supports the hypothesis that architectures optimized through compound scaling effectively capture the multi-scale visual cues characteristic of tornado damage. ResNext variants also perform strongly (e.g., ResNext50-32x4dv2, 60.8\% F1). Perhaps most notably for practical deployment, lightweight architectures achieve top-tier performance while being orders of magnitude smaller: MobileNetV3-Large (62.0\% F1, 4.2M params). \Cref{fig:Figure_9} visualizes the accuracy-efficiency trade-off across all 79 fine-tuned models. A clear Pareto frontier emerges, highlighting the MobileNetV3 family as offering exceptional efficiency, while the EfficientNet variants (particularly B0-B6) and ConvNeXt define the upper bound of accuracy achieved in this stage.
\paragraph{High Ordinal Understanding Achieved, Fine-Grained Precision Still Limited}. Fine-tuning substantially improves performance across all architectures. Most top models achieve Ordinal Top-1 Accuracy $\geq 90$ \% (e.g., EfficientNet-B6: 91.5\%), confirming that fine-tuned networks have learned the relative severity of damage states, aligning with the practical goal of estimating approximate structural damage levels. However, the Macro F1-score, which penalizes even adjacent-class errors, remains notably lower (e.g., EfficientNet-B6: 66.0 \%). This gap reflects the inherent difficulty of distinguishing between intermediate damage states (DS1–DS3), where visual cues overlap and differences are subtle (e.g., roof covering damage below vs. above 15\%, or one vs. two failed windows). While the high Ordinal Accuracy confirms models understand the damage spectrum, the lower Macro F1 highlights the remaining challenge in achieving precise classification for these fine-grained IN-CORE thresholds. This motivates the Stage 2 hyperparameter optimization, which aims to refine discriminative sensitivity for these nuanced intermediate states.

\subsection{Stage 2: Systematic Hyperparameter Optimization}
Building on the Stage 1 baseline, Stage 2 systematically evaluates 29 hyperparameter variations across all 79 architectures, resulting in 2,255 individual training runs. This stage investigates how optimization strategies influence model performance on the QSTD benchmark, with particular emphasis on addressing class imbalance, improving fine-grained discrimination, and enhancing multi-scale feature learning. The analysis focuses on aggregated performance distributions to reveal consistent trends across model families, while detailed per-model results are provided in the Supplementary Material (\Cref{app:detailed_results}).
\subsubsection{Architectural Sensitivity Modulates the Effectiveness of Focal Loss}
To assess whether adaptive re-weighting could improve recognition of under-represented damage states, all 79 architectures were retrained using Focal Loss ($\gamma=5$) and compared to the baseline CCE configuration. The results reveal strong architecture-specific responses rather than a uniform global improvement, highlighting a complex interaction between the loss function's re-weighting mechanism and inherent model properties. 
On average across all models, Focal Loss produced only a marginal change in aggregate metrics (Avg. Macro F1: 52.4\% vs. 52.9\% for CCE; Avg. Ordinal Top-1: 86.1\% vs. 86.2\%). This near-zero average shift, visualized in \Cref{fig:Figure_10}(a), conceals substantial underlying variability: 48.1\% (38/79) of models improved, while 49.4\% (39/79) declined, and 2 (Efficient-B3 and AlexNet) remained unchanged (\Cref{app:stage2_summary}; \Cref{tab:focal_loss_impact}). This indicates that the effectiveness of Focal Loss ($\gamma=5$) in this base configuration is relative.

Model-Level Trends:
\begin{itemize}
    \item \textbf{Residual \& NAS Designs}. Residual architectures generally demonstrated the most consistent improvements. RegNet (9/10 improved), ResNeXt (4/5 improved), and ResNet (5/10 improved) variants appeared robust to the loss re-weighting, suggesting their structural properties (skip connections and balanced design spaces) effectively leverage the emphasized gradients from harder examples.
	\item  \textbf{Transformers (ViT/Swin/MaxViT)}. Transformers showed high variance. ViTs recorded the two largest improvements in our study, ViT-L/16 ($+7.5$ points), while Focal Loss reduced performance in most Swin variants (3 of 6) and the MaxViT baseline (-1.0 point). This highlights a high sensitivity, suggesting potential benefits require careful co-optimization.
	\item  \textbf{ConvNeXt and VGG}. These architectures consistently underperformed with Focal Loss. All four ConvNeXt variants declined significantly (up to $-17.7$ points for ConvNeXt-small), as did most VGG models (6/8). This negative interaction may reflect optimization instability or conflicts between the gradient modulation of Focal Loss and the normalization or propagation mechanisms specific to these designs.
	\item \textbf{EfficientNet \& Lightweight Models}. Showed mixed or negative results. The EfficientNet family (6/12 improved, 5 declined, 1 unchanged) and MobileNet family (1/4 improved) did not show a clear benefit, suggesting their already compact or scaled designs did not universally gain from this specific re-weighting.
\end{itemize}

\begin{figure}[H]
    \centering
    \includegraphics[width=1\linewidth]{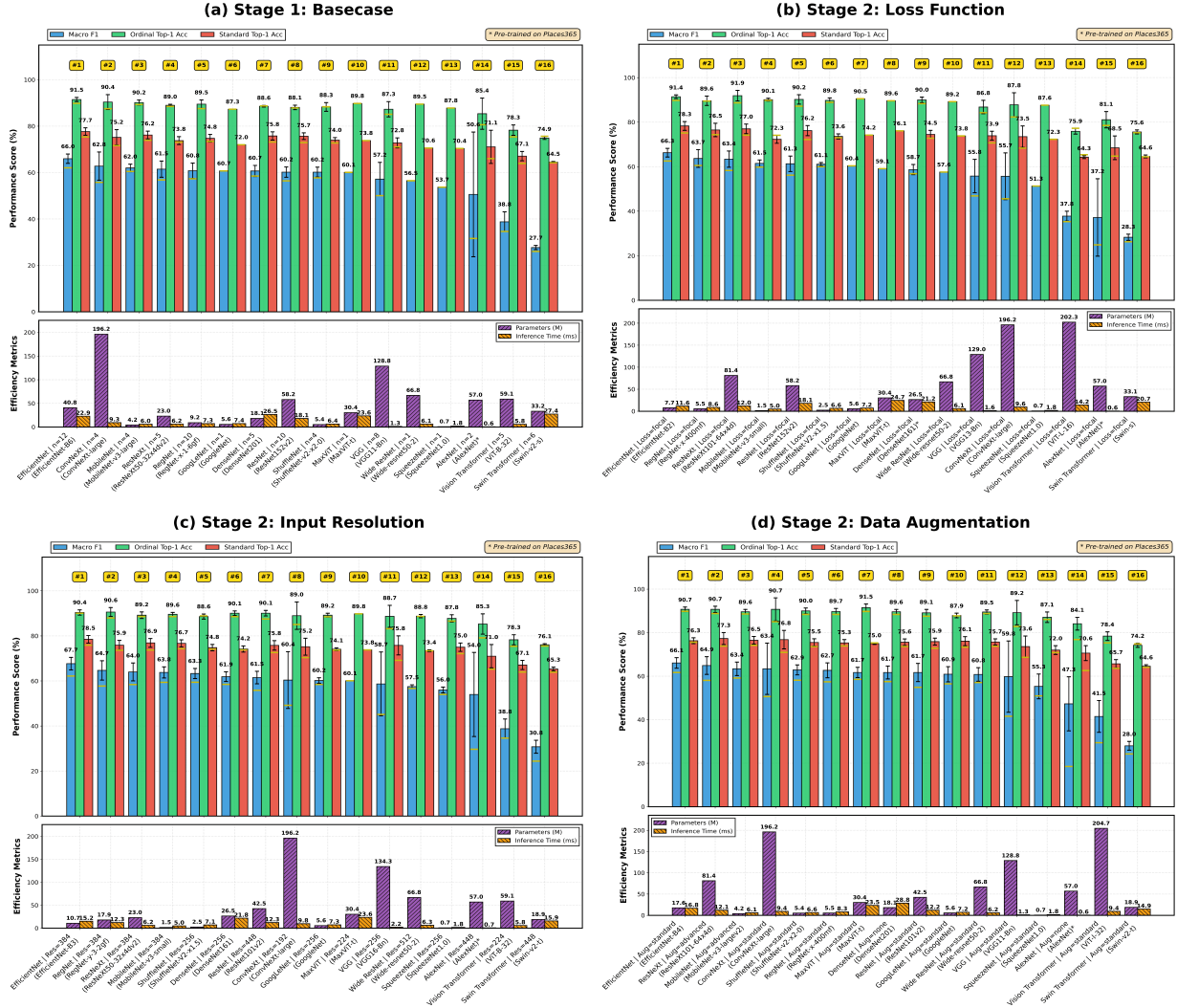}
    \caption{Performance and Efficiency of Top Variants per Family Across Key Experimental Stages. Each subplot shows the best-performing variant from each architectural family for a specific configuration: (a) Stage 1 Base Case ($\psi_{base}$), (b) Stage 2 Optimal Loss Function, (c) Stage 2 Optimal Input Resolution, and (d) Stage 2 Optimal Data Augmentation. (Top Row) Performance comparison using Macro F1 (blue), Ordinal Top-1 Acc (green), and Standard Top-1 Acc (red). (Bottom Row) Corresponding efficiency metrics: Parameter count (purple) and Inference Time (orange)}
    \label{fig:Figure_7}
\end{figure}

\paragraph{Class-Level Impact.} Averaged across all models (\Cref{app:stage2_summary}; \Cref{tab:loss_avg_performance}; \Cref{tab:focal_loss_impact}), Focal Loss did not yield consistent gains for minority damage states compared to CCE in this baseline configuration. Average F1 scores decreased slightly for DS1 ($-1.6$ pts), DS3 ($-0.1$ pts), and DS4 ($-1.1$ pts), with only DS2 showing a minor gain (+0.6 pts). This redistribution aligns with Focal Loss’s bias toward difficult samples but also underscores the limited separability among adjacent classes in QSTD.
Overall, Focal Loss ($\gamma=5$) is not universally beneficial for QSTD when applied in the standard baseline configuration. Its effectiveness is architecture-dependent, improving performance for robust families (RegNet, ResNeXt) while degrading it for others (ConvNeXt, VGG) (\Cref{fig:Figure_7}(b)). The absence of consistent average improvement, especially on minority classes, indicates that effectively addressing QSTD’s imbalance may require architecture-specific tuning of the focusing parameter ($\gamma$) or joint optimization with other hyperparameters. These findings highlight the complex relationship between loss function design, architectural bias, and the intrinsic ordinal ambiguity of multiclass post-disaster damage assessment.

\subsubsection{Optimal Resolution Range on QSTD Emerges Around $256^2$ to $448^2$}
\begin{figure}[htbp]
    \centering
    \includegraphics[width=1\linewidth]{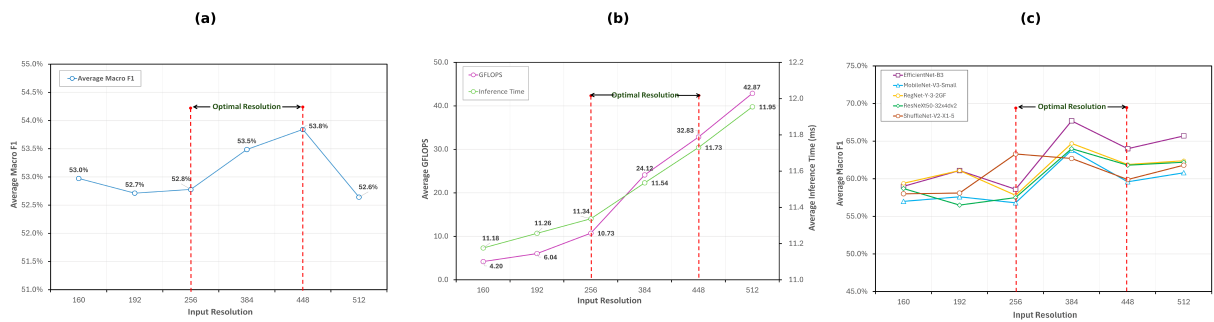}
    \caption{Effect of Input Resolution on Model Performance and Cost. (a) Average Macro F1 score across 73 convolutional model families. (b) Average computational cost (GFLOPs) and inference time (ms), showing a sharp increase at higher resolutions. (c) Performance scaling for representative top-5 performing models, highlighting architecture-specific peaks}
    \label{fig:Figure_11}
\end{figure}

Input resolution is critical for the QSTD dataset, as the IN-CORE taxonomy relies on visual cues that are both subtle and spatially localized. To examine this, we varied input size from 160x160 to 512x512 across all 73 convolutional models. ViT and MaxViT variants were evaluated only at the 224x224 baseline due to patch-size constraints, and batch sizes were reduced for the largest models at high resolutions for memory management (as detailed in \Cref{tab:zero_shot}). 
\paragraph{Optimal Range.} Averaged across all 73 convolutional models, performance improves with increasing resolution, rising from 53.0\% Macro F1 at $160^2$ to a peak of 53.8\% at $448^2$, with a slight decline at $512^2$ (\Cref{fig:Figure_11}a and \Cref{fig:Figure_10}b). This establishes a clear optimal range between $256^2$ and $448^2$, where increased spatial detail contributes meaningfully to damage-state discrimination. Beyond this range, average improvements become marginal while computational costs increase substantially (\Cref{fig:Figure_11}b).
This trend is reinforced by architecture-specific peaks (\Cref{fig:Figure_11}c). Eleven of the sixteen architectural families achieve their highest Macro F1 within the $256^2$ to $448^2$ range (\Cref{fig:Figure_7}c). For example, the top-performing EfficientNet-B3 peaked at $384^2$ (67.7\% Macro F1), as did RegNet-Y-3.2GF (64.7\%) and MobileNet-V3-Small (63.8\%). While exceptions exist (e.g., ConvNeXt-Large peaking early at $192^2$), the strong clustering of optimal performance in the mid-to-high resolution range is consistent.

\paragraph{Class-level effects.} Per-class analysis (\Cref{tab:resolution_avg_performance}, \Cref{fig:qstd_performance_progression_on_image_resolution} and \Cref{fig:heatmap_per_f1}) confirms that the primary beneficiaries of increased resolution are the intermediate damage states. Average F1 scores for the difficult DS1-DS3 classes improved by up to 14.2\% within the optimal range, compared to marginal gains ($\leq$ 1.6\%) for the more easily distinguished DS0, DS4, and DS5 classes. This strongly supports the interpretation that higher resolution is essential for resolving the fine-grained IN-CORE thresholds. 
These findings show that input resolutions between $256\times256$ and $448\times448$ offer the best trade-off between visual fidelity, discriminative performance, and computational efficiency for QSTD. This range captures the necessary detail for fine-grained assessment while remaining practical for deployment.

\subsubsection{Impact of Data Augmentation on Fine-Grained Damage Recognition}
Evaluating data augmentation strategies revealed that moderate transformations consistently enhance model robustness, while overly aggressive approaches degrade performance (\Cref{fig:Figure_10}c). We compared ‘None’, 'Standard’ (moderate geometric transforms), ‘Advanced’ (adding color/perspective shifts, erasing), and ‘Heavy’ (intensified Advanced) strategies against the Basic flip/resize baseline used in Stage 1 (\Cref{app:augmentation_details}).

\begin{figure}[htbp]
    \centering
    \includegraphics[width=1\linewidth]{Figure_10.png}
    \caption{Hyperparameter Sensitivity Analysis Across Architectures. Box plots show the distribution of Macro F1 scores for all 79 fine-tuned models under varying configurations for nine key hyperparameters (*Base Case configuration): (Top row, L-R) Loss Function, Input Resolution, Data Augmentation; (Middle row, L-R) Optimizer, Learning Rate, LR Scheduler; (Bottom row, L-R) Dropout Regularization, Batch Size, Activation Function.  }
    \label{fig:Figure_10}
\end{figure}

\paragraph{Optimal Performance with Standard Augmentation.} As shown by the average performance trends (\Cref{tab:augmentation_avg_performance}), the Standard augmentation strategy achieved the highest average Macro F1 score (54.0\%). This represents a meaningful +2.4 point gain over no augmentation (None: 51.6\%) and a +1.1 point gain over the Basic baseline (52.9\%). These gains were concentrated in the challenging intermediate damage states (DS1–DS3), indicating that moderate geometric variation enhances the model’s ability to recognize subtle structural deterioration. 
\paragraph{Architecture Optimums.} Analysis of the top-performing model within each architectural family confirms that Standard augmentation is the most broadly effective strategy, producing the best results for the majority (12 of 16 families), including ConvNeXt (63.4\% F1) and ResNet (61.7\% F1). A smaller group of robust architectures achieved their peak performance with the Advanced strategy, notably EfficientNet (66.1\% F1) and ResNeXt (64.9\% F1), suggesting they can leverage more complex transformations. Conversely, architectures highly sensitive to spatial or feature perturbation, such as DenseNet and AlexNet, performed best with None.
\paragraph{Detrimental Effect of Aggressive Augmentation.} The Heavy strategy consistently underperformed, yielding the lowest average Macro F1 (46.8\%), a substantial -7.2 points relative to Standard. Critically, Heavy was not the optimal strategy for any of the 16 architectural families. This confirms that aggressive distortions tend to obscure the fine-grained visual cues required by the IN-CORE damage taxonomy, rather than improving robustness.
Consequently, these results demonstrate that Standard augmentation offers the most reliable and broadly effective strategy for QSTD, striking an optimal balance between enhancing variation and preserving structural detail. While certain robust architectures benefit from Advanced transformations, excessive augmentation introduces detrimental noise, particularly for the subtle distinctions required by the IN-CORE damage taxonomy.

\subsubsection{Impact of Optimizer and Learning Rate on Performance}
\begin{figure}[H]
    \centering
    \includegraphics[width=1\linewidth]{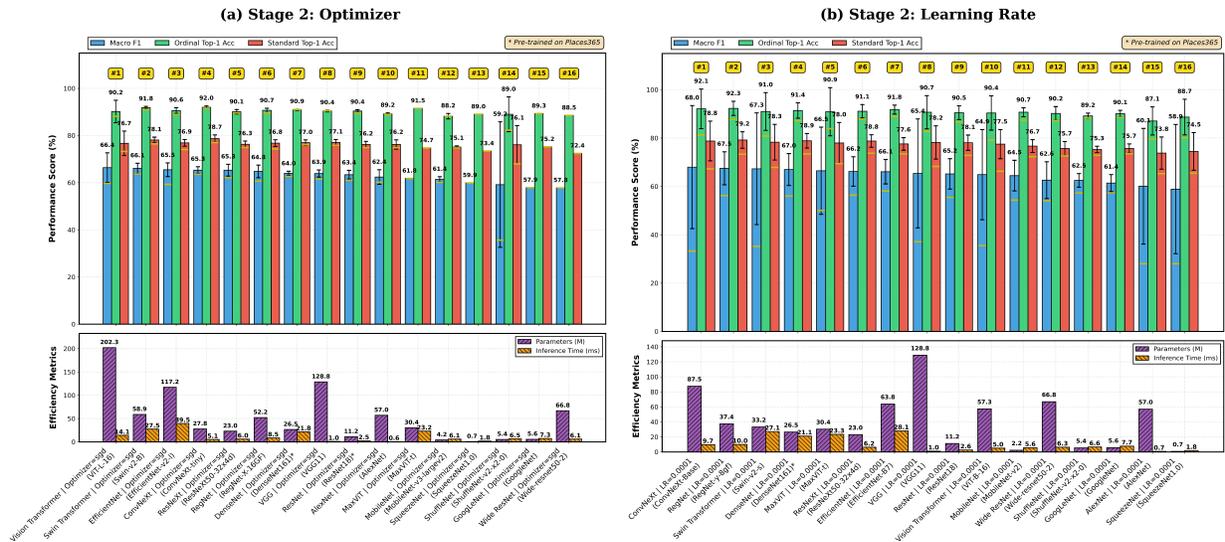}
    \caption{Performance and Efficiency of Top Variants per Family Across Optimizer and Learning Rate. Each subplot shows the best-performing variant from each architectural family for a specific configuration}
    \label{fig:Figure_12}
\end{figure}

While loss function, input resolution, and data augmentation produced broad performance shifts, our analysis revealed deeper interactions between training dynamics and architectural behavior. Among these, the choice of optimizer and learning rate emerged as the most decisive factors influencing model generalization. 
\paragraph{Optimizer Selection Substantially Impacts Transformer Performance.} A central finding of Stage 2 is the strong and differential effect of optimizer choice, particularly for transformers (\Cref{fig:Figure_12}a). Whereas the baseline Adam optimizer delivered stable results for most CNNs in Stage 1, replacing it with Stochastic Gradient Descent (SGD) with momentum resulting in a pronounced and consistent performance boost. Across all 79 models, the average Macro F1 increased from 52.9\% (Adam) to 59.8\% (SGD), accompanied by parallel gains in both standard and ordinal accuracy (\Cref{fig:Figure_10}d and \Cref{tab:optimizer_avg_performance}).
The improvement was especially striking for transformer families. Under Adam, Vision Transformers (ViT) and Swin Transformers ranked near the bottom($\#15$ and $\#16$) of all sixteen model families tested, averaging only 34.6\% and 26.0\% Macro F1, respectively (\Cref{fig:Figure_12}a). When trained with SGD, these families exhibited dramatic jumps: ViT average F1 improved to 59.7\% (+25.1 points), with ViT-L/16 achieving 66.4\% F1, the highest(\#1 rank) of all models in this experiment. Similarly, the Swin Transformer family rose to 63.7\% (+37.7 points), led by Swin-V2-B (66.1\% F1), ranking second overall. These results overturn the Stage 1 observation that transformers lag convolutional models, revealing that the earlier performance gap was primarily driven by optimizer selection rather than architectural limitation.
SGD also benefited most CNN families, though to a lesser extent. ConvNeXt improved by an average of +7.8 points, VGG by +11.5 points, and AlexNet by +28.6 points, yielding top models such as ConvNeXt-Tiny (65.3\% F1). These results suggest that SGD’s momentum-based updates provide better generalization on QSTD’s data, likely by favoring smoother minima. The only major exception was ShuffleNet, whose lightweight architecture showed reduced performance under SGD.
\paragraph{Learning Rate Sensitivity: Stable Optimization at \textbf{$1\times10^{-4}$} is Critical for QSTD.} The learning rate proved to be one of the most sensitive hyperparameters, exerting a strong and consistent influence across architectures. Varying the learning rate from $1\times10^{-4}$ to $5\times10^{-3}$ revealed a clear and pronounced optimum centered at $1\times10^{-4}$, as visualized in \Cref{fig:Figure_10}e. At this setting, the average Macro F1 score across all models reached 63.1\%, a gain of +10.2 points over the Stage 1 baseline value of 52.9\% (at $1\times10^{-3}$). Standard and Ordinal Top-1 accuracy improved similarly to 76.2\% and 90.3\%, respectively (\Cref{fig:Figure_10}e and \Cref{tab:lr_avg_performance}).
As the learning rate increased beyond $1\times10^{-3}$, performance declined sharply. At $3\times10^{-3}$ , average Macro F1 dropped to 43.6\%, and at $5\times10^{-3}$ to 40.9\%, with the most pronounced losses occurring in the intermediate damage states (DS1–DS3). In contrast, the lower learning rate of $1\times10^{-4}$ significantly improved class-level discrimination. Relative to the baseline, average F1 scores increased by +11.7 points for DS1, +10.3 for DS2, +10.5 for DS3, and +17.2 for DS4. This indicates that stable, slow-variance updates are essential for preserving the subtle visual cues that distinguish adjacent severity levels.
This trend also holds consistently when identifying the top-performing model within each architectural family. In all sixteen families, the best-performing variant used $1\times10^{-4}$. Notable examples include the champion model, ConvNeXt-Base (68.0\% F1 – highest on all experiments), RegNet-Y-8GF (67.5\% F1), DenseNet161 (67.0\% F1), EfficientNet-B7 (66.1\% F1), and ResNet18 (65.2\% F1), confirming that this learning rate generalizes across model scale, depth, and inductive bias (\Cref{fig:Figure_12}b). Meanwhile, models trained with larger learning rates exhibited instability, reflected in elevated variance, and reduced minority-class recall.

\subsubsection{Effects of Scheduler, Batch Size, Dropout, and Activation}
\begin{figure}[htbp]
    \centering
    \includegraphics[width=0.95\linewidth]{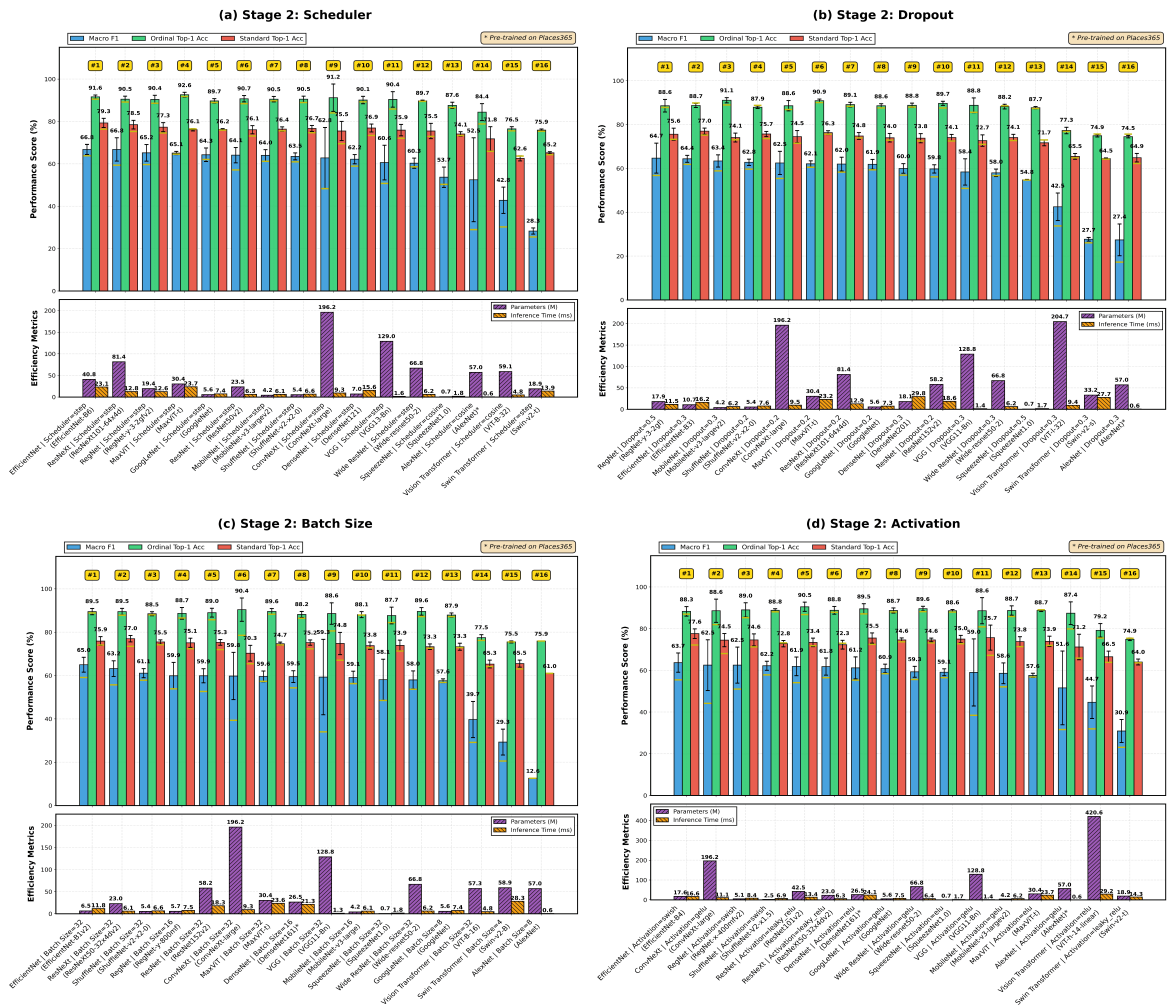}
    \caption{Performance and Efficiency of Top Variants per Family Across Scheduler, Batch Size, Dropout, and Activation Function. Each subplot shows the best-performing variant from each architectural family for a specific configuration}
    \label{fig:Figure_13}
\end{figure}

Further investigation into the remaining hyperparameters revealed moderate or architecture-specific influences, in contrast to the decisive impact of the optimizer and learning rate. The detailed performance distributions for these experiments are visualized in \Cref{fig:Figure_10}(f – i) and \Cref{fig:Figure_13}.
\paragraph{Learning Rate Scheduling.} Introducing a Step decay schedule produced a small but consistent average improvement over a static learning rate (None). Average Macro F1 increased slightly from 52.9\% (None) to 53.8\% (Step), while Cosine scheduling showed negligible difference (52.8\%) (\Cref{fig:Figure_10}f and \Cref{tab:lr_avg_performance}). The Step scheduler yielded the optimal result for the top model in 13 of 16 families, indicating that dynamically reducing the learning rate aids convergence stability, though the absolute gains were modest (\Cref{fig:Figure_13}a).
\paragraph{Dropout.} Applying dropout regularization ($p=0.2$ to $0.5$) showed negligible average impact on performance compared to the baseline ($p=0.0$). Average Macro F1 scores remained virtually unchanged, fluctuating between 52.1\% and 52.9\% (\Cref{fig:Figure_10}g and \Cref{tab:dropout_avg_performance}). While the optimal dropout rate varied for individual models (\Cref{fig:Figure_13}b), the minimal average change suggests that, under this training configuration, dropout was not a critical factor for improving generalization and functioned primarily as a minor, architecture-specific tuning parameter.
\paragraph{Batch Size (BS).} Reducing the batch size from the baseline of 64 generally led to decreased average performance. Average Macro F1 steadily dropped from 52.9\% (BS=64) to 51.0\% (BS=32), 48.5\% (BS=16), and sharply declined at BS=8 (45.7\%) and BS=4 (43.3\%). This trend, particularly the significant drop below BS=16, suggests that with the Adam optimizer, the benefits of stable gradients from larger batches outweigh potential noise-induced regularization, possibly exacerbated by unreliable Batch Normalization statistics at very small sizes (\Cref{fig:Figure_13}(c) and \Cref{tab:batch_size_avg_performance}).

\paragraph{Activation Function.} To ensure a fair comparison, we evaluated standardized implementations of these functions search space (ELU, Leaky ReLU, Swish/SiLU, GeLU, ReLU) across all models, even when they matched the architecture’s native default. Replacing native activations (e.g., SiLU, ReLU, GeLU) with these explicit alternatives resulted in lower average performance (\Cref{fig:Figure_13}d). The default setting (Avg. F1: 52.9\%) outperformed all alternatives, which ranges from 44.5\% to 50.5\% (\Cref{fig:Figure_10}h and \Cref{tab:activation_avg_performance}). This confirms that the default architecture-specific activations are generally well-optimized, and forcing a change provides no consistent benefit on average.

\subsubsection{Pre-Training Source Analysis: ImageNet vs. Places365}
\begin{figure}[H]
    \centering
    \includegraphics[width=1\linewidth]{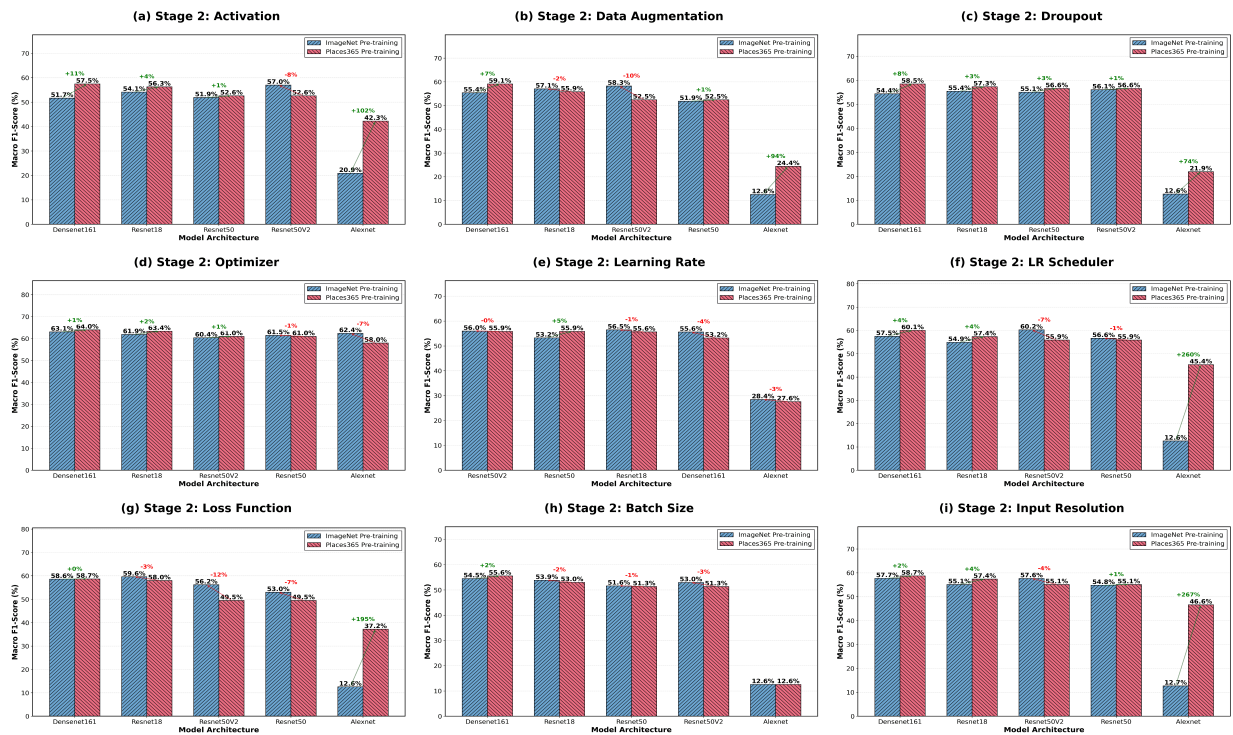}
    \caption{Pre-training Source Comparison: Average Macro F1 Across Stage 2 Optimizations (Places365 vs. ImageNet-1K)}
    \label{fig:Figure_14}
\end{figure}

The choice between object-centric (ImageNet) and scene-centric (Places365) pre-training significantly influenced performance, revealing architecture-dependent effects that evolved across experimental stages. Places365 demonstrated strong initial transferability, offering dramatic zero-shot F1 gains for AlexNet (+301.6\%) and DenseNet161 (+142.3\%). This advantage persisted after baseline CCE fine-tuning (Stage 1) for AlexNet, DenseNet161 (+3.8\%), and ResNet18 (+7.8\%), highlighting the benefit of scene-context features under standard conditions.
However, Stage 2 optimization introduced complex interactions, as visualized in \Cref{fig:Figure_14}. While Places365 variants maintained an average advantage across the dropout experiments (5/5 pairs) and under SGD optimization for specific models (3/5 pairs), ImageNet proved superior under other key conditions. Notably, the optimal low learning rate ($1\times10^{-4}$) favored ImageNet variants on average (4/5 pairs), as did Focal Loss for the ResNet family. This suggests that while Places365 aids initial adaptation, ImageNet's object-centric features may be more robust or responsive to the specific optimization techniques needed to enhance fine-grained discrimination. Ultimately, neither pre-training source was universally optimal across all conditions.

\subsection{Stage 3: Zero-Shot Generalization Test on the TMTD Dataset}
\begin{figure}[H]
    \centering
    \includegraphics[width=1\linewidth]{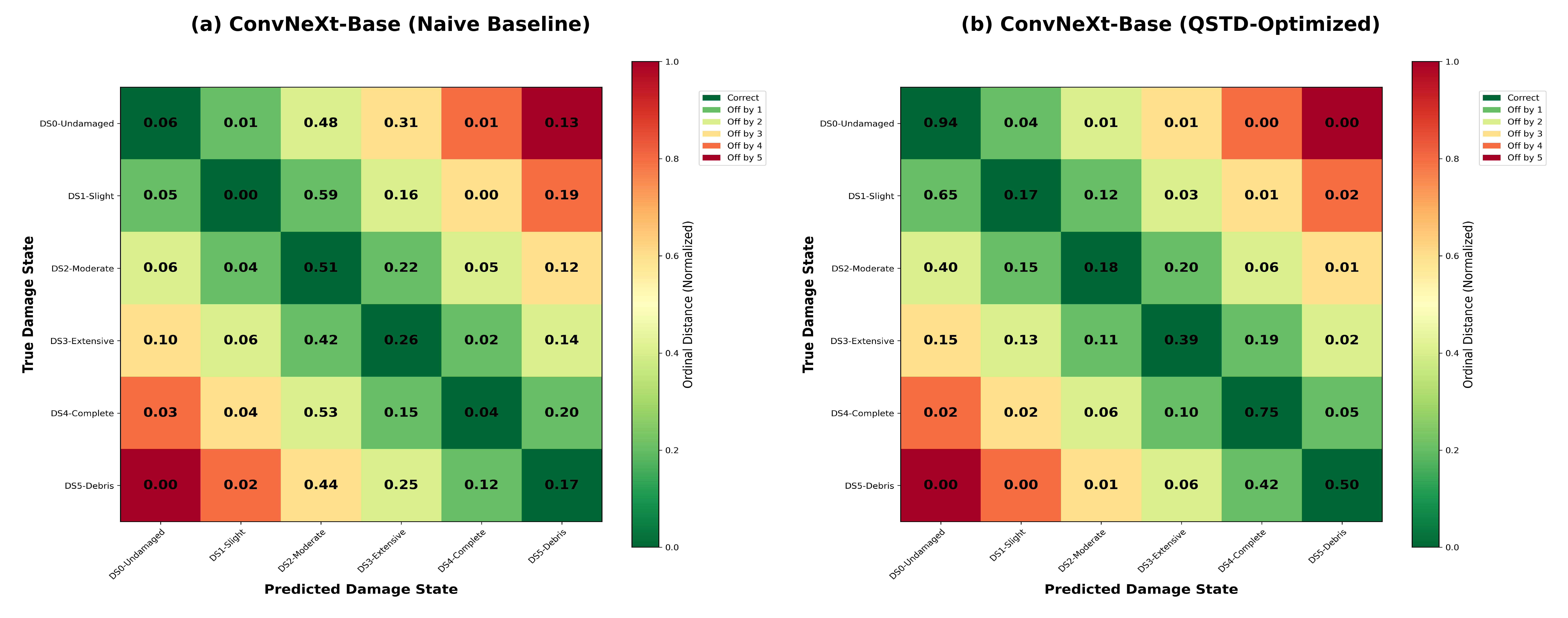}
    \caption{Ordinal Confusion Matrices for Zero-Shot TMTD Test. (a) Naïve baseline model showing a poor classification performance. (b) QSTD-optimized model showing generalizable ordinal understanding of damage severity on an unseen dataset}
    \label{fig:Figure_15}
\end{figure}

To assess real-world robustness under domain shift, we evaluated the optimized champion model (ConvNeXt-Base, fine-tuned on QSTD with learning rate $1\times10^{-4}$) in a zero-shot setting on the TMTD dataset. We compared its performance to a naive baseline: the same architecture with its original ImageNet pre-training applied directly to TMTD without QSTD fine-tuning.
As shown in \Cref{tab:tmtd_generalization}, the naive baseline failed to transfer, achieving only 11.8\% Macro F1 and 43.9\% Ordinal Accuracy, with predictions collapsing toward intermediate classes. In sharp contrast, our QSTD-trained model achieved 46.4\% Macro F1 (a +34.6-point improvement) and retained an exceptionally high 85.5\% Ordinal Top-1 Accuracy, indicating preserved damage ordering even when fine-grained discrimination was difficult.
The confusion matrices in \Cref{fig:Figure_15} visually confirm this finding. The naive baseline (a) exhibits near-random errors and classification collapse. In contrast, our QSTD-optimized model (b) shows errors tightly clustered along the main and adjacent diagonals, correctly identifying DS0 and DS4 while preserving the ordinal structure for intermediate states. This demonstrates that while the model struggles with fine-grained distinctions due to the domain shift, our QSTD-based framework successfully imparted a robust, generalizable ordinal understanding of damage severity that was entirely absent in the baseline model.

\begin{table*}[htbp]
\centering
\caption{Zero-Shot Generalization Performance on the TMTD Dataset (ConvNeXt-Base).}
\label{tab:tmtd_generalization}
\small
\renewcommand{\arraystretch}{1.3}
\begin{tabular}{@{}lcccccccccc@{}}
\toprule
\textbf{Model Configuration} & \textbf{Macro F1} & \textbf{Ordinal} & \textbf{Std.} & \textbf{DS0} & \textbf{DS1} & \textbf{DS2} & \textbf{DS3} & \textbf{DS4} & \textbf{DS5} \\
& \textbf{(\%)} & \textbf{Top-1 Acc (\%)} & \textbf{Top-1 Acc (\%)} & \textbf{(F1)} & \textbf{(F1)} & \textbf{(F1)} & \textbf{(F1)} & \textbf{(F1)} & \textbf{(F1)} \\
\midrule
Naive Baseline & \multirow{2}{*}{11.8} & \multirow{2}{*}{43.9} & \multirow{2}{*}{14.0} & \multirow{2}{*}{9.3} & \multirow{2}{*}{0.7} & \multirow{2}{*}{19.4} & \multirow{2}{*}{17.8} & \multirow{2}{*}{7.9} & \multirow{2}{*}{15.9} \\
(ImageNet $\rightarrow$ TMTD) & & & & & & & & & \\
% \midrule
QSTD-Optimized & \multirow{2}{*}{\textbf{46.4}} & \multirow{2}{*}{\textbf{85.5}} & \multirow{2}{*}{\textbf{58.3}} & \multirow{2}{*}{\textbf{63.5}} & \multirow{2}{*}{\textbf{20.6}} & \multirow{2}{*}{\textbf{22.6}} & \multirow{2}{*}{\textbf{36.2}} & \multirow{2}{*}{\textbf{75.5}} & \multirow{2}{*}{\textbf{59.1}} \\
(ImageNet $\rightarrow$ QSTD $\rightarrow$ TMTD) & & & & & & & & & \\
\bottomrule
\end{tabular}
\end{table*}

\begin{table}[H]
\centering
\caption{Practical Optimization Guidelines based on QSTD Benchmark (Applicable Across All 79 Architectures).}
\label{tab:optimization_guidelines}
\small
\setlength{\tabcolsep}{3pt}
\renewcommand{\arraystretch}{1.3}
\begin{tabular}{@{}p{3cm}p{4cm}p{4cm}p{4.5cm}@{}}
\toprule
\textbf{Component} & \textbf{Recommendation} & \textbf{Impact} & \textbf{Notes} \\
\midrule
Optimizer & SGD with momentum $=$ 0.9 for Vision Transformers; Adam acceptable for CNNs & +25 to +38 F1 points for ViT/Swin families & Most impactful optimization; elevates Transformers from bottom-tier to competitive performance \\
\midrule
Learning Rate & $1 \times 10^{-4}$ (universal across all families) & +10.2 F1 points on average & Rates $\geq 1 \times 10^{-3}$ consistently resulted in poorer performance \\
\midrule
Input Resolution & $256 \times 256$ to $448 \times 448$ & +14.2\% F1 for DS1--DS3 & Higher resolutions show diminishing returns with increased cost \\
\midrule
Data Augmentation & Standard (moderate geometric transforms) & +2.4 F1 vs. none; $-7.2$ F1 with heavy augmentation & Aggressive augmentation degrades fine-grained damage discrimination \\
\midrule
Loss Function & Architecture-dependent tuning & Mixed effects (48\% improved; 49\% degraded with Focal Loss) & Cross-entropy remains the most robust baseline on the QSTD benchmark \\
\midrule
Pre-training Source & ImageNet (default); Places365 for AlexNet, DenseNet161, ResNet18 & Up to +142\% F1 (zero-shot) & Scene-centric pre-training benefits certain architectures \\
\midrule
Cross-Event Generalization & Expect $>85\%$ ordinal accuracy; 20--30 F1 point drop & 85.5\% ordinal accuracy on TMTD & Damage severity ranking generalizes better than fine-grained class labels \\
\bottomrule
\end{tabular}
\end{table}

\begin{table}[H]
\centering
\caption{QSTD-Based Deployment Configurations for Automated Tornado Damage Assessment.}
\label{tab:deployment_configurations}
\small
\setlength{\tabcolsep}{4pt}
\renewcommand{\arraystretch}{1.3}
\begin{tabular}{@{}p{2.0cm}p{2.1cm}p{1.8cm}p{1.3cm}p{1.5cm}p{1.3cm}p{2.0cm}p{2.5cm}@{}}
\toprule
\textbf{Deployment} & \textbf{Recommended} & \textbf{Optimizer} & \textbf{Learning} & \textbf{Input} & \textbf{Aug-} & \textbf{Performance} & \textbf{Key} \\
\textbf{Scenario} & \textbf{Model} & & \textbf{Rate} & \textbf{Resolution} & \textbf{mentation} & \textbf{Metrics} & \textbf{Advantages} \\
\midrule
Resource-Constrained (Edge devices, mobile platforms) & MobileNetV3-Large or ShuffleNetV2-x1.5 & SGD (momentum=0.9) & $1\times10^{-4}$ & $256\times256$ & Standard & 62.0\% F1, 4.2M params, $\sim$5ms inference & Optimal accuracy-efficiency trade-off for limited compute \\
\midrule
Maximum Accuracy (GPU-enabled backend) & ConvNeXt-Base & SGD (momentum=0.9) & $1\times10^{-4}$ & $384\times384$ & Standard & 68.0\% F1 (QSTD), 46.4\% F1 (TMTD), 85.5\% Ordinal Acc (TMTD) & Highest accuracy with strong cross-event generalization \\
\bottomrule
\end{tabular}
\end{table}

\section{Discussions and Recommendations}\label{GSA}
Our systematic largescale evaluation across 79 architectures yields actionable guidelines for deploying automated tornado damage assessment systems based on QSTD benchmark. The Pareto frontier analysis from Stage 1 base case experiments (\Cref{fig:Figure_9}) shows that the MobileNetV3 family offers exceptional computational efficiency while maintaining competitive accuracy, whereas EfficientNet (B0–B6) and ConvNeXt variants define the upper bound of achievable performance. Building on these findings, \Cref{tab:optimization_guidelines} presents practical optimization guidelines validated across all architectural families, and \Cref{tab:deployment_configurations} summarizes deployment configurations optimized for distinct computational constraints.

\section{Conclusions}
This paper introduced a large-scale, systematic experimental framework to deconstruct the architectural and optimization drivers of post-tornado damage recognition, leveraging 79 open-source models in over 2,300 controlled experiments on the QSTD benchmark. Our findings provide a comprehensive performance baseline and practical guidance for developing robust, open-source damage assessment tools.
Our key findings show that performance is contingent on the complex interaction between architecture, pretraining source, and optimization, rather than on architectural choice alone. We established the following:
\begin{itemize}
	\item \textbf{Domain-specific fine-tuning is essential.} Zero-shot models failed to generalize regardless of scale or design. Pre-training on scene-centric Places365 provided a significant initial transfer advantage for architectures like AlexNet ($+301.6$\% F1) and DenseNet ($+142.3$\% F1) over object-centric ImageNet weights, though this benefit varied once deeper optimization strategies were introduced in later stages.
	\item \textbf{Optimizer choice can be more critical than architecture.} While CNNs outperformed Transformers under Adam during baseline fine-tuning, switching to SGD with momentum produced large performance gains for ViT and Swin Transformer families ($+25$ to $+38$ points F1), moving them from among the lowest-ranked to competitive with top-tier CNNs.
	\item \textbf{EfficientNet variants consistently emerge as top-tier performers.} The EfficientNet family secured a dominant share of the best-performing configurations, including 7 of the top 10 ranks across all Stage 2 optimization experiments. Its balanced scaling design proved highly effective for this multi-scale damage task, with EfficientNet-B6 achieving the highest Stage 1 baseline Macro F1 score (66.0\%).
	\item \textbf{Key hyperparameters demonstrated distinct and non-obvious effects.} A low learning rate of $1\times 10^{-4}$ was universally critical for high performance yielding an average Macro F1 of 63.1\%, a $+10.2$ point gain over the $1\times 10^{-3}$ baseline. Input resolution and data augmentation both exhibited optimal "sweet spots" (peaking at approximately $256^2$ to $448^2$ and 'Standard' augmentation, respectively), with overly aggressive settings proving detrimental. In contrast, Focal Loss, dropout, and batch size variations produced architecture-dependent outcomes rather than consistent, universal improvement.
	\item \textbf{The framework successfully imparts ordinal understanding.} In the Stage 3 zero-shot generalization test on the TMTD dataset, the QSTD-optimized champion model (ConvNeXt-Base) achieved a 46.4\% Macro F1 score, improving by $+34.6$ points over the naive baseline ($11.8$\%), and retained an 85.5\% Ordinal Top-1 Accuracy. This indicates that the model learned the relative structure of damage severity even under domain shift.
\end{itemize}

These results are promising; however, some identified opportunities remain that provide clear paths for future investigation. The current reliance on ground-level 2D imagery captures only the damage visible from a single viewpoint, suggesting that future studies could incorporate multi-view, UAV, or LiDAR data to better address occlusions. Additionally, while our experiments varied one factor at a time to enable clear performance attribution, future work could explore higher-order parameter interactions through automated search or joint optimization. Finally, the drop in performance during transfer testing also highlights the challenge of cross-event generalization, which may be ensured through lightweight domain adaptation or few-shot calibration. Collectively, this study quantifies the specific drivers of performance for post-tornado damage recognition and demonstrates that effective optimization requires the careful co-adaptation of architecture, optimizer, and hyperparameters. It establishes a robust foundation for transitioning automated street-view assessment from a research prototype into a dependable operational capability, providing emergency responders with the rapid, objective intelligence required to enhance community resilience and save lives in the wake of catastrophic events.

\paragraph{Acknowledgments.}
The authors acknowledge the Center for Risk-Based Community Resilience Planning, a NIST-funded Center of Excellence (Cooperative Agreement 70NANB15H044), for providing the street-view imagery dataset used in this study. This research was made possible through the data resources shared by the Center, which were instrumental in the development and validation of the models presented in this work.

\newpage

\appendix
\crefname{section}{Appendix}{Appendices}
\Crefname{section}{Appendix}{Appendices}
\setcounter{figure}{0}
\setcounter{table}{0}
\renewcommand{\thefigure}{A\arabic{figure}}
\renewcommand{\thetable}{A\arabic{table}}
{\centering \Large \textbf{Appendix} \par}
\vspace{0.2em}
% \addcontentsline{toc}{section}{Appendix}

\section{Dataset Details}
\label{app:dataset_details}

\subsection{QSTD Annotation and Curation Protocol}
\label{app:annotation_protocol}

This section details the two-phase protocol for dataset creation.

\subsubsection{Phase 1: Independent Annotation}

Two trained annotators, each with background in structural engineering and disaster assessment, independently processed the complete set of 4,800 source images. Their task involved:

\begin{itemize}
    \item \textbf{Building Identification}: Identifying all residential structures visible in each frame that exhibited sufficient clarity for damage assessment.
    
    \item \textbf{Crop Extraction}: Manually cropping individual buildings to create focused, building-centric images while excluding confounding background elements.
    
    \item \textbf{Damage Classification}: Assigning damage state labels (DS0--DS5) based on visible structural component failures according to the \INCORE{} criteria defined in \Cref{tab:damage_taxonomy}.
\end{itemize}

This initial phase yielded 5,583 labeled building crops. The discrepancy between 4,800 source frames and 5,583 crops (ratio of 1.16:1) reflects the presence of multiple assessable buildings per frame in densely damaged neighborhoods.

\subsubsection{Phase 2: Cross-Validation and Consensus Building}

To ensure dataset quality and labeling consistency, the annotators conducted a comprehensive cross-validation review of each other's work. This process involved:

\begin{itemize}
    \item \textbf{Crop Quality Verification}: Ensuring building boundaries were accurately delineated and that crops contained sufficient structural details for damage assessment.
    
    \item \textbf{Duplicate Detection}: Identifying instances where the same building was cropped from multiple overlapping frames.
    
    \item \textbf{Label Consistency Review}: Evaluating damage state assignments for adherence to \INCORE{} criteria and consistency with similar cases.
\end{itemize}

The cross-validation phase revealed several systematic patterns requiring resolution:

\begin{itemize}
    \item \textbf{Image Quality Issues}: Sixty-six images (1.2\%) were flagged for quality concerns including excessive motion blur, severe occlusion ($>50\%$ of structure obscured), or lighting conditions preventing reliable assessment of key structural elements. These images were excluded from the final dataset.
    
    \item \textbf{Duplicate Buildings}: Multiple instances of the same building captured from different angles or in overlapping frames were identified. In these cases, the annotators selected the crop with optimal viewing angle and clarity while excluding redundant instances.
    
    \item \textbf{Boundary Case Classifications}: The most substantial reclassification occurred at the DS2/DS3 boundary, particularly regarding the interpretation of roof sheathing failure extent. Initial conservative labeling was adjusted during consensus review to strictly align with the \INCORE{} threshold criterion ($>3$ failed sections for DS3). This refinement process resulted in 241 images being reclassified from DS2 to DS3, reflecting the challenge of consistently quantifying partial structural failures in 2D imagery.
\end{itemize}

\begin{figure}
    \centering
    \includegraphics[width=1\linewidth]{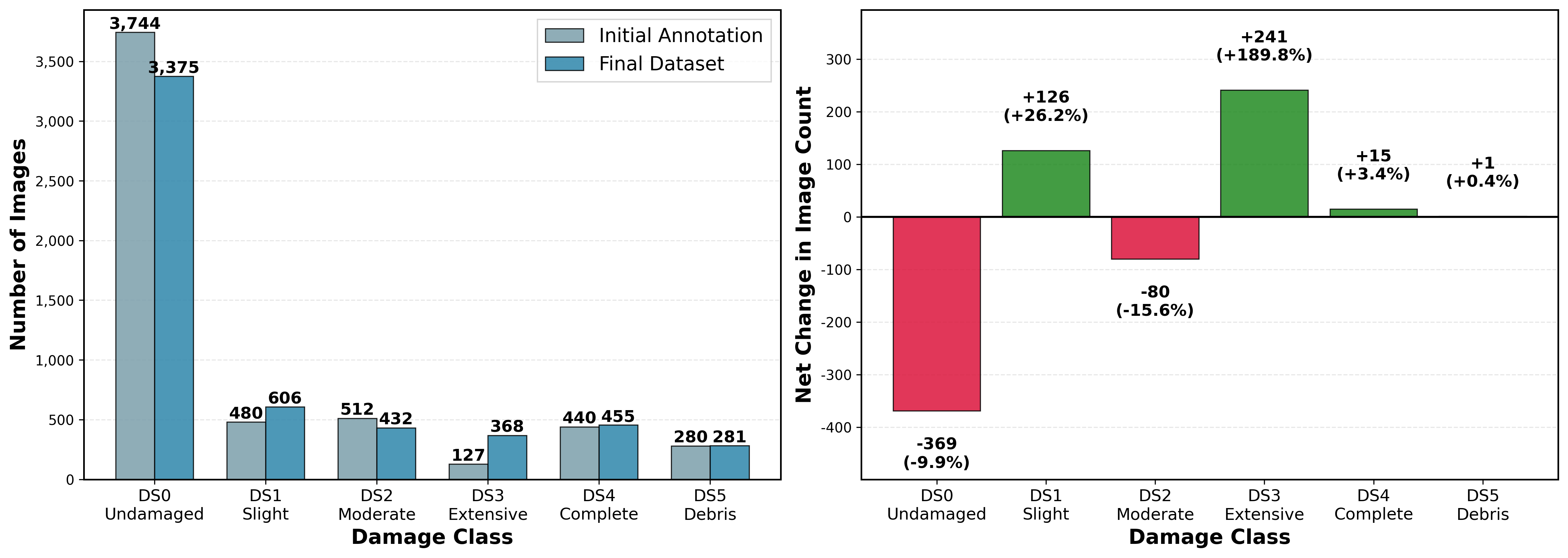}
    \caption{QSTD Dataset Evolution and Refinement through Inter-Annotator Cross-Validation}
    \label{fig:dataset_evolution}
\end{figure}

\Cref{tab:class_evolution} and \Cref{fig:dataset_evolution} summarize the evolution and primary adjustments of class distribution through the curation process:

\begin{table}[htbp]
\centering
\caption{Class Distribution Changes During Annotation and Cross-Validation.}
\label{tab:class_evolution}
\small
\setlength{\tabcolsep}{5pt}  % default is 6pt
\renewcommand{\arraystretch}{1.2}
\begin{tabular}{@{}lcccp{6cm}@{}}
\toprule
\textbf{Class} & \textbf{Initial} & \textbf{Final} & \textbf{Net Change} & \textbf{Primary Adjustment} \\
& \textbf{Annotation} & \textbf{Dataset} & & \\
\midrule
DS0 - Undamaged & 3,744 & 3,375 & $-369$ ($-9.9\%$) & Reclassified to DS1 upon detection of subtle roof/window damage \\
% \midrule
DS1 - Slight & 480 & 606 & $+126$ ($+26.3\%$) & Received reclassifications from DS0 and DS2 \\
% \midrule
DS2 - Moderate & 512 & 432 & $-80$ ($-15.6\%$) & Many reclassified to DS3 based on roof sheathing criteria \\
% \midrule
DS3 - Extensive & 127 & 368 & $+241$ ($+189.8\%$) & Significant increase due to consistent application of sheathing failure thresholds \\
% \midrule
DS4 - Complete & 440 & 455 & $+15$ ($+3.4\%$) & Minor adjustments; clearest damage state with least ambiguity \\
% \midrule
DS5 - Debris/Non-Structural & 280 & 281 & $+1$ ($+0.4\%$) & Stable; most objective classification \\
\midrule
\textbf{Total} & \textbf{5,583} & \textbf{5,517} & \textbf{$-66$ ($-1.2\%$)} & \textbf{Quality-based exclusions} \\
\bottomrule
\end{tabular}
\end{table}

\subsection{QSTD Class and Resolution Statistics}
\label{app:resolution_stats}

This section provides supplementary data on the final \QSTD{} dataset.

\begin{table}[htbp]
\centering
\caption{QSTD Image Dimensions by Damage Class. The table lists the mean width and height in pixels for images within each of the six damage state categories after initial cropping.}
\label{tab:image_dimensions}
\small
\setlength{\tabcolsep}{18pt} 
\renewcommand{\arraystretch}{1.8}
\begin{tabular}{@{}lcc@{}}
\toprule
\textbf{Damage Class} & \textbf{Average Width (px)} & \textbf{Average Height (px)} \\
\midrule
DS0 - Undamaged & 347.02 & 225.70 \\
DS1 - Slight & 371.34 & 238.19 \\
DS2 - Moderate & 367.34 & 229.52 \\
DS3 - Extensive & 366.38 & 235.88 \\
DS4 - Complete & 373.15 & 248.80 \\
DS5 - Debris & 1048.47 & 632.23 \\
\bottomrule
\end{tabular}
\end{table}

\begin{figure}
    \centering
    \includegraphics[width=1\linewidth]{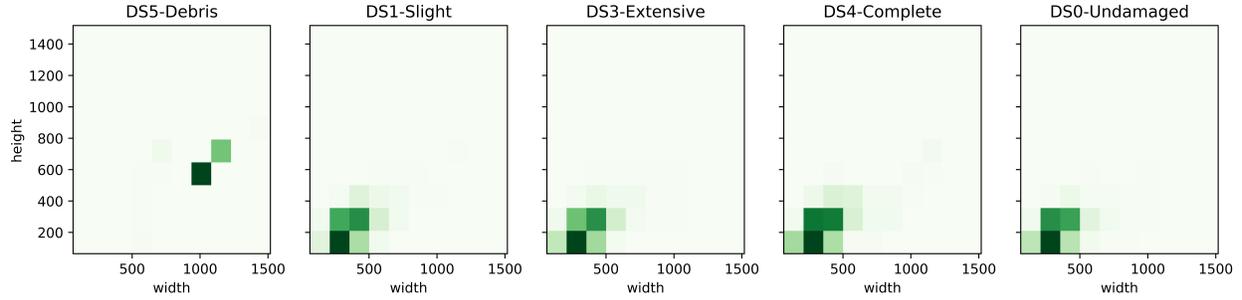}
    \caption{Figure showing the distribution of QSTD image resolutions (Height x Width) for each of the 6 damage classes, highlighting the different characteristics of the DS5 class}
    \label{fig:qstd_image_resolution}
\end{figure}

%%% NEW SECTION
\setcounter{figure}{0}
\setcounter{table}{0}
\renewcommand{\thefigure}{B\arabic{figure}}
\renewcommand{\thetable}{B\arabic{table}}

\section{Experimental Framework Details}
\label{app:experimental_framework}

\subsection{Full Definition of Data Augmentation Strategy}
\label{app:augmentation_details}

This section explicitly defines the data augmentation strategy used in the study, grouped by their types.

\begin{table*}[htbp]
\centering
\caption{Comprehensive List of Data Augmentation Strategies Used in Stage 1 and 2.}
\label{tab:augmentation_full}
\small
\setlength{\tabcolsep}{5pt}  % default is 6pt
\renewcommand{\arraystretch}{1.32}
\begin{tabular}{@{}lp{4cm}p{4.1cm}p{5cm}@{}}
\toprule
\textbf{Strategy} & \textbf{Transformations} & \textbf{Key Parameters} & \textbf{Context} \\
\midrule
None & -- & -- & Provides a non-augmented baseline; used for validation and test sets. \\
\midrule
Basic & Horizontal flip & $p = 0.3$ & Basic invariance to viewpoint. \\
(Base Case) & Resize & Target input resolution & \\
\midrule
Standard & Horizontal flip & $p = 0.5$ & Simulates moderate changes in camera angle and framing \\
& Random rotation & $\theta \sim \mathcal{U}(-15°, 15°)$ & \\
& Random crop & From padded input & \\
\midrule
Advanced & Horizontal flip & $p = 0.5$ & Simulates more significant viewpoint changes, lighting variations, and minor occlusions. \\
& Vertical flip & $p = 0.2$ & \\
& Random rotation & $\theta \sim \mathcal{U}(-30°, 30°)$ & \\
& Affine transformation & $t \sim \mathcal{U}(-0.1, 0.1)$ & \\
& Perspective distortion & $\sigma = 0.3$, $p = 0.3$ & \\
& Color jittering & $\pm 0.2$ (brightness, contrast, saturation), $\pm 0.1$ (hue) & \\
& Random erasing & $p = 0.3$, scale $\sim \mathcal{U}(0.02, 0.25)$ & \\
\midrule
Heavy & All advanced transforms with intensified parameters: & & Tests the upper limit of model robustness against extreme variations and occlusions. \\
& Random rotation & $\theta \sim \mathcal{U}(-45°, 45°)$ & \\
& Affine transformation & $t \sim \mathcal{U}(-0.15, 0.15)$, scale $\sim \mathcal{U}(0.85, 1.15)$ & \\
& Perspective distortion & $\sigma = 0.4$, $p = 0.4$ & \\
& Color jittering & $\pm 0.3$ (brightness, contrast, saturation), $\pm 0.15$ (hue) & \\
& Random erasing & $p = 0.5$ & \\
\bottomrule
\end{tabular}
\end{table*}

\subsection{Full List of 79 Evaluated Model Variants}
\label{app:model_list}

This section explicitly lists all model variants used in the study, grouped by their 16 architectural families.

\begin{table*}[htbp]
\centering
\caption{Comprehensive List of All 79 Model Variants.}
\label{tab:complete_models}
\small
\renewcommand{\arraystretch}{1.6}
\begin{tabular}{@{}clcp{11cm}@{}}
\toprule
\textbf{\#} & \textbf{Architecture} & \textbf{Number of} & \textbf{Models} \\
& & \textbf{Variants} & \\
\midrule
1 & AlexNet & 2 & AlexNet, AlexNet-Places365 \\
% \midrule
2 & ConvNeXt & 4 & ConvNeXt-Tiny, ConvNeXt-Small, ConvNeXt-Base, ConvNeXt-Large \\
% \midrule
3 & DenseNet & 5 & DenseNet121, DenseNet161, DenseNet161-Places365, DenseNet169, DenseNet201 \\
% \midrule
4 & EfficientNet & 12 & EfficientNet-B0, EfficientNet-B1, EfficientNet-B1v2*, EfficientNet-B2, EfficientNet-B3, EfficientNet-B4, EfficientNet-B5, EfficientNet-B6, EfficientNet-B7, EfficientNet-v2-s, EfficientNet-v2-m, EfficientNet-v2-l \\
% \midrule
5 & GoogLeNet & 1 & GoogLeNet \\
% \midrule
6 & MaxViT & 1 & MaxViT-t \\
% \midrule
7 & MobileNet & 4 & MobileNet-v2, MobileNet-v3-Small, MobileNet-v3-Large, MobileNet-v3-Largev2* \\
% \midrule
8 & RegNet & 10 & RegNet-x-400mf, RegNet-x-400mfv2*, RegNet-x-1-6gf, RegNet-x-16gf, RegNet-y-800mf, RegNet-y-3-2gf, RegNet-y-3-2gfv2*, RegNet-y-8gf, RegNet-y-32gf, RegNet-y-128gf \\
% \midrule
9 & ResNet & 10 & ResNet18, ResNet18-Places365, ResNet34, ResNet50, ResNet50-Places365, ResNet50v2, ResNet101, ResNet101v2*, ResNet152, ResNet152v2* \\
% \midrule
10 & ResNeXt & 5 & ResNeXt50-32x4d, ResNeXt50-32x4dv2*, ResNeXt101-32x8d, ResNeXt101-32x8dv2*, ResNeXt101-64x4d \\
% \midrule
11 & ShuffleNet & 4 & ShuffleNet-v2-x0-5, ShuffleNet-v2-x1-0, ShuffleNet-v2-x1-5, ShuffleNet-v2-x2-0 \\
% \midrule
12 & SqueezeNet & 1 & SqueezeNet1-0 \\
% \midrule
13 & Swin & 6 & Swin-B, Swin-S, Swin-T, Swin-v2-B, Swin-v2-S, Swin-v2-T \\
% \midrule
14 & VGG & 8 & VGG11, VGG11-BN, VGG13, VGG13-BN, VGG16, VGG16-BN, VGG19, VGG19-BN \\
% \midrule
15 & ViT & 5 & ViT-B-16, ViT-B-32, ViT-L-16, ViT-L-32, ViT-H-14-Linear \\
% \midrule
16 & Wide ResNet & 1 & Wide-ResNet50-2 \\
\midrule
& \textbf{Total} & \textbf{79} & \\
\bottomrule
\multicolumn{4}{l}{\scriptsize * Models were initialized using the more recent IMAGENET1K\_V2 weights}
\end{tabular}
\end{table*}

%NEW SESSION
%%% NEW SECTION
\setcounter{figure}{0}
\setcounter{table}{0}
\renewcommand{\thefigure}{C\arabic{figure}}
\renewcommand{\thetable}{C\arabic{table}}

\section{Detailed Experimental Results}
\label{app:detailed_results}

\subsection{Full Stage 1 (Base Case Fine-Tuned) Results}
\label{app:stage1_results}

Table~\ref{tab:stage1_full} presents the complete performance results for all 79 model variants trained using the $\psi_{\text{base}}$ configuration.

\begin{longtable}{@{}clp{0.5cm}cccccc@{}}
\caption{Complete Performance Results for All 79 Model Variants Trained Using the $\psi_{\text{base}}$ Configuration.}
\setlength{\tabcolsep}{5pt}  % default is 6pt
\label{tab:stage1_full} \\
\toprule
% \normalsize
\textbf{Rank} & \textbf{Model} & \textbf{Pre-training} & \textbf{Macro F1} & \textbf{Ordinal Top-1} & \textbf{Standard Top-1} & \textbf{Params} & \textbf{Inference} \\
& & & \textbf{(Val) (\%)} & \textbf{Acc (Val) (\%)} & \textbf{Acc (Val) (\%)} & \textbf{(M)} & \textbf{Time (ms)} \\
\midrule
\endfirsthead

\multicolumn{8}{c}{\tablename\ \thetable\ -- \textit{Continued from previous page}} \\
\toprule
\textbf{Rank} & \textbf{Model} & \textbf{Pre-training} & \textbf{Macro F1} & \textbf{Ordinal Top-1} & \textbf{Standard Top-1} & \textbf{Params} & \textbf{Inference} \\
& & & \textbf{(Val) (\%)} & \textbf{Acc (Val) (\%)} & \textbf{Acc (Val) (\%)} & \textbf{(M)} & \textbf{Time (ms)} \\
\midrule
\endhead

\midrule
\multicolumn{8}{r}{\textit{Continued on next page}} \\
\endfoot

\bottomrule
\multicolumn{8}{l}{ \quad \quad \quad \quad \quad \quad \quad \quad \quad \quad \quad \quad\quad \quad \quad \quad\quad \quad \quad \scriptsize \dag~ImageNet \quad \quad \quad \quad $\bigstar$~Places365}
\endlastfoot
1 & EfficientNet-B6 & \dag & \textbf{66.0} & \textbf{91.5} & \textbf{77.7} & 40.75 & 22.89 \\
2 & EfficientNet-B3 & \dag & 64.4 & 88.7 & 77.0 & 10.71 & 13.29 \\
3 & EfficientNet-B4 & \dag & 63.9 & 89.8 & 77.4 & 17.56 & 16.65 \\
4 & EfficientNet-B1v2 & \dag & 63.4 & 89.1 & 75.5 & 6.52 & 11.70 \\
5 & ConvNeXt-Large & \dag & 62.8 & 90.4 & 75.2 & 196.21 & 9.34 \\
6 & EfficientNet-B0 & \dag & 62.6 & 90.6 & 76.2 & 4.02 & 8.19 \\
7 & MobileNet-v3-Large & \dag & 62.0 & 90.2 & 76.2 & 4.21 & 6.00 \\
8 & EfficientNet-B7 & \dag & 61.8 & 90.1 & 73.0 & 63.80 & 28.24 \\
9 & EfficientNet-B1 & \dag & 61.6 & 91.0 & 75.9 & 6.52 & 11.84 \\
10 & MobileNet-v3-Small & \dag & 61.6 & 89.2 & 72.2 & 1.52 & 4.96 \\
11 & ResNeXt50-32x4dv2 & \dag & 61.5 & 89.0 & 73.8 & 22.99 & 6.21 \\
12 & RegNet-x-1-6gf & \dag & 60.8 & 89.5 & 74.8 & 9.19 & 7.27 \\
13 & GoogLeNet & \dag & 60.7 & 87.3 & 72.0 & 5.61 & 7.42 \\
14 & EfficientNet-B2 & \dag & 60.7 & 89.3 & 75.5 & 7.71 & 12.15 \\
15 & DenseNet201 & \dag & 60.7 & 88.6 & 75.8 & 18.10 & 26.55 \\
16 & EfficientNet-v2-m & \dag & 60.3 & 90.1 & 75.2 & 52.87 & 26.71 \\
17 & ShuffleNet-v2-x2-0 & \dag & 60.2 & 88.3 & 74.0 & 5.36 & 6.45 \\
18 & ResNet152v2 & \dag & 60.2 & 88.1 & 75.7 & 58.16 & 18.07 \\
19 & MaxViT-t & \dag & 60.1 & 89.8 & 73.8 & 30.38 & 23.61 \\
20 & DenseNet169 & \dag & 60.1 & 88.3 & 71.8 & 12.49 & 21.58 \\
21 & EfficientNet-v2-l & \dag & 60.1 & 90.0 & 73.4 & 117.24 & 37.90 \\
22 & EfficientNet-v2-s & \dag & 60.1 & 88.7 & 73.5 & 20.19 & 19.14 \\
23 & EfficientNet-B5 & \dag & 59.9 & 89.1 & 73.9 & 28.35 & 20.40 \\
24 & MobileNet-v3-Largev2 & \dag & 59.8 & 90.1 & 73.8 & 4.21 & 6.12 \\
25 & DenseNet121 & \dag & 59.7 & 88.8 & 72.7 & 6.96 & 15.43 \\
26 & RegNet-x-400mf & \dag & 59.5 & 88.2 & 70.8 & 5.50 & 8.42 \\
27 & RegNet-x-400mfv2 & \dag & 59.3 & 88.3 & 72.3 & 5.10 & 8.48 \\
28 & ConvNeXt-Base & \dag & 58.9 & 89.8 & 72.9 & 87.55 & 9.40 \\
29 & RegNet-y-800mf & \dag & 58.8 & 87.9 & 74.7 & 6.43 & 7.73 \\
30 & RegNet-y-3-2gfv2 & \dag & 58.7 & 89.5 & 75.9 & 19.44 & 11.77 \\
31 & MobileNet-v2 & \dag & 58.5 & 87.9 & 73.5 & 2.23 & 5.61 \\
32 & ResNeXt101-32x8dv2 & \dag & 58.5 & 88.5 & 72.9 & 86.76 & 12.18 \\
33 & ShuffleNet-v2-x1.5 & \dag & 58.3 & 87.9 & 72.0 & 2.49 & 6.57 \\
34 & RegNet-y-3-2gf & \dag & 58.1 & 87.8 & 72.4 & 19.44 & 11.76 \\
35 & ResNet18 & $\bigstar$ & 57.9 & 89.5 & 73.5 & 11.18 & 2.53 \\
36 & ResNet101v2 & \dag & 57.8 & 89.2 & 74.2 & 42.51 & 12.27 \\
37 & ResNet50v2 & \dag & 57.7 & 88.1 & 73.0 & 23.52 & 6.13 \\
38 & ShuffleNet-v2-x1.0 & \dag & 57.5 & 90.9 & 73.4 & 1.26 & 6.65 \\
39 & ResNeXt101-64x4d & \dag & 57.2 & 88.7 & 74.8 & 81.42 & 11.92 \\
40 & VGG11-BN & \dag & 57.2 & 87.3 & 72.8 & 128.80 & 1.34 \\
41 & RegNet-x-16gf & \dag & 57.1 & 85.4 & 72.8 & 54.28 & 8.58 \\
42 & DenseNet161 & $\bigstar$ & 57.1 & 88.7 & 74.1 & 26.49 & 21.33 \\
43 & VGG13-BN & \dag & 56.8 & 86.7 & 71.7 & 128.98 & 1.55 \\
44 & RegNet-y-8gf & \dag & 56.8 & 88.6 & 75.2 & 39.38 & 9.73 \\
45 & ResNet50 & \dag & 56.7 & 87.1 & 70.2 & 23.52 & 6.31 \\
46 & Wide-ResNet50-2 & \dag & 56.5 & 89.5 & 70.6 & 66.85 & 6.11 \\
47 & ResNet34 & \dag & 56.5 & 86.5 & 72.9 & 21.29 & 4.64 \\
48 & ResNet101 & \dag & 56.4 & 87.2 & 72.2 & 42.51 & 12.25 \\
49 & VGG16-BN & \dag & 55.8 & 87.8 & 73.5 & 134.29 & 1.97 \\
50 & ResNet152 & \dag & 55.0 & 87.3 & 72.6 & 58.16 & 18.01 \\
51 & DenseNet161 & \dag & 55.0 & 89.3 & 71.5 & 26.49 & 21.64 \\
52 & ShuffleNet-v2-x0-5 & \dag & 55.0 & 86.4 & 71.7 & 0.35 & 6.41 \\
53 & RegNet-y-32gf & \dag & 54.9 & 86.7 & 73.8 & 145.05 & 12.12 \\
54 & ResNeXt50-32x4d & \dag & 54.7 & 87.8 & 73.8 & 22.99 & 6.17 \\
55 & ResNet18 & \dag & 53.7 & 88.2 & 72.3 & 11.18 & 2.49 \\
56 & SqueezeNet1.0 & \dag & 53.7 & 87.8 & 70.4 & 0.74 & 1.77 \\
57 & ResNet50 & $\bigstar$ & 52.5 & 87.4 & 72.4 & 23.52 & 6.14 \\
58 & ResNeXt101-32x8d & \dag & 52.4 & 88.5 & 70.3 & 86.76 & 11.89 \\
59 & VGG19-BN & \dag & 51.7 & 87.4 & 71.2 & 139.61 & 2.38 \\
60 & ConvNeXt-Tiny & \dag & 51.4 & 86.4 & 71.0 & 27.82 & 4.91 \\
61 & AlexNet & $\bigstar$ & 50.6 & 85.4 & 71.1 & 57.03 & 0.61 \\
62 & ConvNeXt-Small & \dag & 50.1 & 83.5 & 67.4 & 49.44 & 9.47 \\
63 & VGG13 & \dag & 49.4 & 84.8 & 72.1 & 128.98 & 1.17 \\
64 & RegNet-y-128gf & \dag & 49.2 & 83.8 & 72.0 & 644.81 & 17.78 \\
65 & VGG16 & \dag & 47.8 & 84.8 & 71.6 & 134.29 & 1.45 \\
66 & VGG11 & \dag & 46.1 & 84.5 & 68.7 & 128.79 & 0.98 \\
67 & ViT-B-32 & \dag & 38.8 & 78.3 & 67.1 & 59.07 & 5.85 \\
68 & ViT-H-14-Linear & \dag & 37.6 & 72.3 & 61.9 & 202.34 & 14.21 \\
69 & ViT-L-32 & \dag & 36.7 & 75.4 & 64.4 & 204.70 & 9.37 \\
70 & VGG19 & \dag & 35.6 & 77.7 & 67.0 & 139.60 & 1.73 \\
71 & ViT-L-16 & \dag & 30.3 & 76.5 & 63.8 & 202.34 & 14.10 \\
72 & ViT-B-16 & \dag & 29.7 & 74.5 & 62.6 & 57.30 & 5.14 \\
73 & Swin-v2-S & \dag & 27.7 & 74.9 & 64.5 & 33.23 & 27.41 \\
74 & Swin-S & \dag & 26.2 & 75.9 & 64.4 & 33.05 & 20.95 \\
75 & Swin-v2-B & \dag & 25.9 & 76.5 & 64.1 & 58.95 & 27.18 \\
76 & Swin-T & \dag & 25.8 & 75.3 & 64.5 & 18.86 & 10.56 \\
77 & Swin-v2-T & \dag & 25.5 & 76.1 & 63.8 & 18.94 & 14.29 \\
78 & Swin-B & \dag & 25.1 & 75.4 & 64.0 & 58.72 & 21.81 \\
79 & AlexNet & \dag & 12.6 & 75.9 & 61.0 & 57.03 & 0.62 \\
\midrule
& ImageNet Avg & \dag & 52.8 & 86.1 & 71.7 & 60.33 & 11.35 \\
& Places365 Avg & $\bigstar$ & 54.5 & 87.7 & 72.8 & 29.55 & 7.65 \\
% \bottomrule
% \multicolumn{8}{l}{\scriptsize \dag~ImageNet \quad $\bigstar$~Places365}
\end{longtable}

\subsection{Stage 2 (Hyperparameter Optimization) Summary Data}
\label{app:stage2_summary}
This section provides the aggregated data used to generate the summary plots in the main paper.
\subsubsection{Loss Function}

\begin{table}[H]
\centering
\caption{Average Performance for Loss Function: CCE vs. Focal Loss ($\gamma=5$, $\alpha=0.5$).}
\label{tab:loss_avg_performance}
\setlength{\tabcolsep}{35pt}  % default is 6pt
\small
\renewcommand{\arraystretch}{1.25}
\begin{tabular}{@{}lccc@{}}
\toprule
\textbf{Metrics (Average)} & \textbf{CCE*} & \textbf{Focal Loss} & \textbf{$\Delta$} \\
\midrule
Macro F1 & 52.9 & 52.4 & $-0.5$ \\
Standard Top-1 & 71.7 & 71.8 & 0.0 \\
Ordinal Top-1 & 86.2 & 86.1 & $-0.1$ \\
Params (M) & 58.77 & 61.54 & $+2.77$ \\
GFLOPs & 8.98 & 9.85 & $+0.86$ \\
Inference Time (ms) & 11.16 & 11.42 & $+0.26$ \\
\midrule
DS0 (F1) & 84.6 & 84.3 & $-0.3$ \\
DS1 (F1) & 20.2 & 18.6 & $+0.6$ \\
DS2 (F1) & 20.2 & 20.9 & $+0.7$ \\
DS3 (F1) & 41.3 & 41.1 & $-0.1$ \\
DS4 (F1) & 63.1 & 62.0 & $-1.1$ \\
DS5 (F1) & 88.1 & 87.7 & $-0.4$ \\
\bottomrule
\multicolumn{4}{l}{\scriptsize * Base Case Configuration}
\end{tabular}
\end{table}

\begin{table}[H]
\centering
\caption{Per-Family Impact of Focal Loss ($\gamma=5$, $\alpha=0.5$) vs. CCE Baseline.}
\label{tab:focal_loss_impact}
\small
\renewcommand{\arraystretch}{1.6}
\setlength{\tabcolsep}{1.4pt}
\begin{tabular}{@{}lccccc@{}}
\toprule
\textbf{Architectural Family} & \textbf{Variants} & \textbf{Variants} & \textbf{Avg. F1 Change} & \textbf{Top F1} & \textbf{Top F1} \\
& \textbf{Tested} & \textbf{Improved (\%)} & \textbf{(Points)} & \textbf{Gain} & \textbf{Loss} \\
\midrule
AlexNet & 2 & 0 (0) & $-6.70$ & --- & $-13.4$ (AlexNet-Places365) \\
ConvNeXt & 4 & 0 (0) & $-10.20$ & --- & $-17.7$ (ConvNeXt-Base) \\
DenseNet & 5 & 2 (40) & $-1.70$ & $+1.6$ (DenseNet161) & $-6.0$ (DenseNet121) \\
EfficientNet & 12 & 6 (50) & $+0.50$ & $+5.6$ (EfficientNet-B2) & $-3.4$ (EfficientNet-B6) \\
GoogLeNet & 1 & 0 (0) & $-0.30$ & --- & $-0.3$ (GoogLeNet) \\
MaxViT & 1 & 0 (0) & $-1.00$ & --- & $-1.0$ (MaxViT-t) \\
MobileNet & 4 & 1 (25) & $-0.70$ & $+0.8$ (MobileNet-v3-Largev2) & $-3.3$ (MobileNet-v3-Large) \\
RegNet & 10 & 9 (90) & $+2.90$ & $+5.4$ (RegNet-Y-16GF) & $-0.9$ (RegNet-Y-3-2GF) \\
ResNet & 10 & 5 (50) & $-0.20$ & $+5.9$ (ResNet18) & $-3.7$ (ResNet50) \\
ResNeXt & 5 & 4 (80) & $+1.40$ & $+6.2$ (ResNeXt101-64x4d) & $-2.1$ (ResNeXt50-32x4dv2) \\
ShuffleNet & 4 & 3 (75) & $+2.20$ & $+4.4$ (ShuffleNet-v2-x0-5) & $-0.5$ (ShuffleNet-v2-x2-0) \\
SqueezeNet & 1 & 0 (0) & $-2.40$ & --- & $-2.4$ (SqueezeNet-1-0) \\
Swin Transformer & 6 & 3 (50) & $+0.30$ & $+2.1$ (Swin-S) & $-1.7$ (Swin-T) \\
VGG & 8 & 2 (25) & $-2.10$ & $+2.1$ (VGG19) & $-14.4$ (VGG13) \\
Vision Transformer & 5 & 2 (40) & $+0.70$ & $+7.5$ (ViT-L-16) & $-3.0$ (ViT-L-32) \\
Wide ResNet & 1 & 1 (100) & $+1.10$ & $+1.1$ (Wide-ResNet50-2) & --- \\
\midrule
\textbf{Total} & \textbf{79} & \textbf{38 (48\%)} & & & \\
\bottomrule
\end{tabular}
\end{table}

\subsubsection{Input Resolution}

\begin{table}[H]
\centering
\caption{Average Macro F1 and Efficiency Metrics Across All Models for Input Resolution.}
\label{tab:resolution_avg_performance}
\small
\setlength{\tabcolsep}{9pt}
\renewcommand{\arraystretch}{1.5}
\begin{tabular}{@{}lcccccccc@{}}
\toprule
\textbf{Metrics (Average)} & \textbf{224*} & \textbf{160} & \textbf{192} & \textbf{256} & \textbf{384} & \textbf{448} & \textbf{512} \\
\midrule
Macro F1 & 52.9\% & 53.0\% & 52.7\% & 52.8\% & 53.5\% & \textbf{53.9\%} & 52.7\% \\
Standard Top-1 & 71.7\% & 72.1\% & 71.8\% & \textbf{72.4\%} & \textbf{72.4\%} & 72.3\% & 72.1\% \\
Ordinal Top-1 & 86.2\% & 86.4\% & 86.4\% & 86.5\% & 86.5\% & \textbf{86.9\%} & 86.1\% \\
Params (M) & 58.77 & 52.57 & 52.57 & 52.57 & 52.57 & 52.57 & 52.57 \\
GFLOPs & 8.98 & 4.17 & 6.00 & 10.66 & 23.96 & 32.61 & 42.59 \\
Inference Time (ms) & 11.16 & 11.11 & 11.19 & 11.27 & 11.46 & 11.66 & 11.88 \\
\midrule
DS0 (F1) & \textbf{92.4} & 84.8 & 84.6 & 85.0 & 85.1 & 85.0 & 84.9 \\
DS1 (F1) & 17.8 & 21.2 & 20.7 & 20.4 & \textbf{23.4} & 22.1 & 22.0 \\
DS2 (F1) & 16.9 & 20.3 & 19.6 & 20.4 & \textbf{21.4} & 19.4 & 19.8 \\
DS3 (F1) & 37.1 & 42.3 & 42.3 & 41.4 & 42.0 & \textbf{45.2} & 41.6 \\
DS4 (F1) & 62.3 & 63.2 & 62.4 & 62.7 & 63.0 & \textbf{64.1} & 60.4 \\
DS5 (F1) & 52.9 & 86.0 & 86.6 & 87.0 & 86.2 & \textbf{87.5} & 87.2 \\
\bottomrule
\multicolumn{8}{l}{\scriptsize * Base Case Configuration}
\end{tabular}
\end{table}

\begin{figure}
    \centering
    \includegraphics[width=1\linewidth]{Figure_C1.png}
    \caption{Figure showing the performance progression of QSTD image resolutions  (Height x Width) for each of the 6 damage classes.}
    \label{fig:qstd_performance_progression_on_image_resolution}
\end{figure}

\begin{figure}
    \centering
    \includegraphics[width=0.75\linewidth]{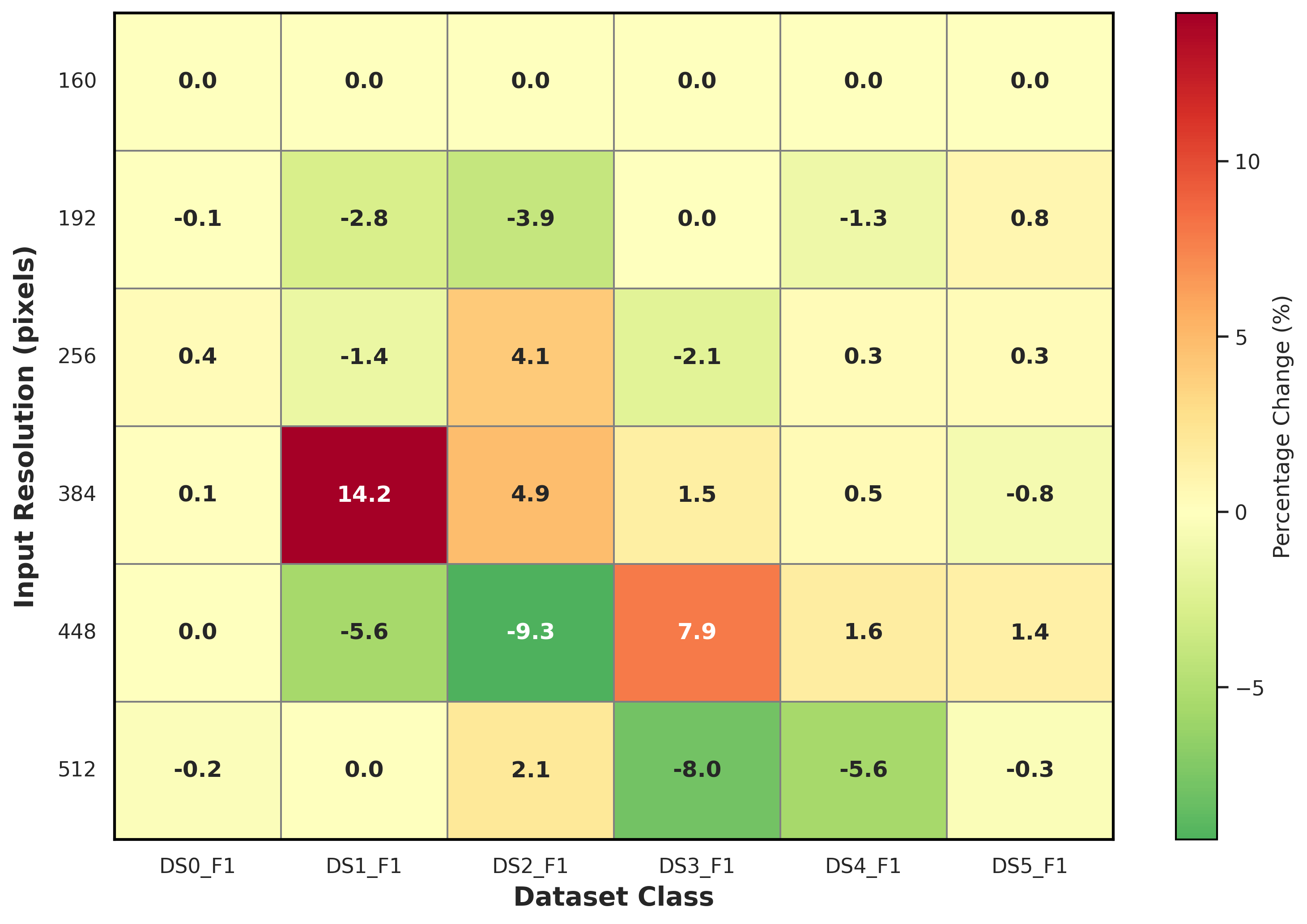}
    \caption{Heatmap of Per-Class F1 Score Percentage Change by Input Resolution, Relative to the $224^2$ Baseline. The blue-bounded region ($256^2$ – $448^2$) highlights the optimal resolution range, where the most significant performance gains (+14.2 for DS1) are concentrated in the challenging intermediate damage states (DS1-DS3).}
    \label{fig:heatmap_per_f1}
\end{figure}

\subsubsection{Data Augmentation}

\begin{table}[H]
\centering
\caption{Average Macro F1 and Efficiency Metrics Across All Models for Data Augmentation Strategies.}
\label{tab:augmentation_avg_performance}
\setlength{\tabcolsep}{13pt}
\small
\renewcommand{\arraystretch}{1.5}
\begin{tabular}{@{}lccccc@{}}
\toprule
\textbf{Augmentation Strategy} & \textbf{None} & \textbf{Basic*} & \textbf{Standard} & \textbf{Advanced} & \textbf{Heavy} \\
\midrule
Avg. Macro F1 & 51.6\% & 52.9\% & \textbf{54.0\%} & 50.2\% & 46.8\% \\
Avg. Standard Top-1 & 71.3\% & 71.7\% & \textbf{72.5\%} & 71.9\% & 70.7\% \\
Avg. Ordinal Top-1 & 85.7\% & 86.2\% & \textbf{86.4\%} & 85.8\% & 84.6\% \\
Avg. Params (M) & 58.78 & 58.77 & 58.77 & 58.78 & 61.54 \\
Avg. GFLOPs & 8.98 & 8.98 & 8.98 & 8.98 & 9.85 \\
Avg. Inference Time (ms) & 11.23 & 11.16 & 11.23 & 11.21 & 11.43 \\
\midrule
DS0 (F1) & 84.4 & 84.6 & \textbf{84.9} & 84.5 & 84.1 \\
DS1 (F1) & 19.8 & 20.2 & \textbf{22.7} & 20.0 & 15.0 \\
DS2 (F1) & 18.0 & 20.2 & \textbf{22.1} & 17.8 & 12.9 \\
DS3 (F1) & 40.2 & 41.3 & \textbf{43.8} & 39.8 & 37.8 \\
DS4 (F1) & 60.4 & \textbf{63.1} & 62.6 & 57.4 & 51.3 \\
DS5 (F1) & 86.0 & \textbf{88.1} & 87.9 & 81.7 & 79.5 \\
\bottomrule
\multicolumn{6}{l}{\scriptsize * Base Case Configuration}
\end{tabular}
\end{table}

\subsubsection{Optimizer}

\begin{table}[H]
\centering
\caption{Average Performance for Optimizer (Adam vs. SGD).}
\label{tab:optimizer_avg_performance}
\small
\setlength{\tabcolsep}{35pt}
\renewcommand{\arraystretch}{1.5}
\begin{tabular}{@{}lccc@{}}
\toprule
\textbf{Metrics (Average)} & \textbf{Adam*} & \textbf{SGD} & \textbf{$\Delta$} \\
\midrule
Macro F1 & 52.9\% & \textbf{59.8\%} & $+6.9\%$ \\
Standard Top-1 & 71.7\% & \textbf{74.6\%} & $+2.9\%$ \\
Ordinal Top-1 & 86.2\% & \textbf{89.3\%} & $+3.2\%$ \\
Params (M) & 58.77 & 61.28 & $+2.51$ \\
GFLOPs & 8.98 & 9.85 & $+0.86$ \\
Inference Time (ms) & 11.16 & 11.42 & $+0.26$ \\
\midrule
DS0 (F1) & 84.6 & \textbf{86.6} & $+2.0$ \\
DS1 (F1) & 20.2 & \textbf{28.2} & $+8.0$ \\
DS2 (F1) & 20.2 & \textbf{25.3} & $+5.1$ \\
DS3 (F1) & 41.3 & \textbf{51.8} & $+10.5$ \\
DS4 (F1) & 63.1 & \textbf{76.0} & $+12.9$ \\
DS5 (F1) & 88.1 & \textbf{90.8} & $+2.7$ \\
\bottomrule
\multicolumn{4}{l}{\scriptsize * Base Case Configuration}
\end{tabular}
\end{table}

\subsubsection{Learning Rate}

\begin{table}[H]
\centering
\caption{Average Performance for Learning Rate Configurations.}
\label{tab:lr_avg_performance}
\small
\setlength{\tabcolsep}{17pt}
\renewcommand{\arraystretch}{1.5}
\begin{tabular}{@{}lcccc@{}}
\toprule
\textbf{Metrics (Average)} & \textbf{$1\times10^{-3}$*} & \textbf{$1\times10^{-4}$} & \textbf{$3\times10^{-3}$} & \textbf{$5\times10^{-3}$} \\
\midrule
Macro F1 & 52.9\% & \textbf{63.1\%} & 43.6\% & 40.9\% \\
Standard Top-1 & 71.7\% & \textbf{76.2\%} & 69.1\% & 68.3\% \\
Ordinal Top-1 & 86.2\% & \textbf{90.3\%} & 83.8\% & 82.5\% \\
Params (M) & 58.77 & 61.28 & 61.28 & 61.28 \\
GFLOPs & 8.98 & 9.85 & 9.85 & 9.85 \\
Inference Time (ms) & 11.16 & 11.43 & 11.40 & 11.41 \\
\midrule
DS0 (F1) & 92.4 & 92.6 & 94.5 & \textbf{94.8} \\
DS1 (F1) & 17.8 & \textbf{28.6} & 10.8 & 8.7 \\
DS2 (F1) & 16.9 & \textbf{25.2} & 10.4 & 9.1 \\
DS3 (F1) & 37.1 & \textbf{48.9} & 28.9 & 28.0 \\
DS4 (F1) & 62.3 & \textbf{79.4} & 47.5 & 43.5 \\
DS5 (F1) & 52.9 & \textbf{63.1} & 43.6 & 40.9 \\
\bottomrule
\multicolumn{5}{l}{\scriptsize * Base Case Configuration}
\end{tabular}
\end{table}

\subsubsection{Learning Rate Scheduler}

\begin{table}[H]
\centering
\caption{Average Performance for LR Scheduler Strategies.}
\label{tab:scheduler_avg_performance}
\small
\setlength{\tabcolsep}{35pt}
\renewcommand{\arraystretch}{1.6}
\begin{tabular}{@{}lccc@{}}
\toprule
\textbf{Metrics (Average)} & \textbf{None*} & \textbf{Cosine} & \textbf{Step} \\
\midrule
Macro F1 & 52.9\% & 52.8\% & \textbf{53.8\%} \\
Standard Top-1 & 71.7\% & 71.8\% & \textbf{73.2\%} \\
Ordinal Top-1 & 86.2\% & 86.0\% & \textbf{86.9\%} \\
Params (M) & 58.77 & 61.53 & 61.53 \\
GFLOPs & 8.98 & 9.85 & 9.85 \\
Inference Time (ms) & 11.16 & 11.46 & 11.39 \\
\midrule
DS0 (F1) & 92.4 & 92.5 & \textbf{94.3} \\
DS1 (F1) & 17.8 & 17.8 & \textbf{18.5} \\
DS2 (F1) & 16.9 & 15.8 & \textbf{17.8} \\
DS3 (F1) & 37.1 & \textbf{38.9} & 37.5 \\
DS4 (F1) & 62.3 & 62.2 & \textbf{63.0} \\
DS5 (F1) & 52.9 & 52.8 & \textbf{53.8} \\
\bottomrule
\multicolumn{4}{l}{\scriptsize * Base Case Configuration}
\end{tabular}
\end{table}

\subsubsection{Batch Size}

\begin{table}[H]
\centering
\caption{Average Performance for Batch Size Configurations.}
\label{tab:batch_size_avg_performance}
\small
\renewcommand{\arraystretch}{1.4}
\setlength{\tabcolsep}{15pt}
\begin{tabular}{@{}lccccc@{}}
\toprule
\textbf{Metrics (Average)} & \textbf{64*} & \textbf{32} & \textbf{16} & \textbf{8} & \textbf{4} \\
\midrule
Macro F1 & \textbf{52.9\%} & 51.0\% & 48.5\% & 45.7\% & 43.3\% \\
Standard Top-1 & \textbf{71.7\%} & 71.4\% & 70.8\% & 70.1\% & 69.2\% \\
Ordinal Top-1 & \textbf{86.2\%} & 85.4\% & 84.8\% & 84.1\% & 82.8\% \\
Params (M) & 58.77 & 61.28 & 61.28 & 61.28 & 61.28 \\
GFLOPs & 8.98 & 9.85 & 9.85 & 9.85 & 9.85 \\
Inference Time (ms) & 11.16 & 11.40 & 11.37 & 11.41 & 11.37 \\
\midrule
DS0 (F1) & \textbf{92.4} & 84.3 & 83.8 & 83.3 & 82.9 \\
DS1 (F1) & 17.8 & \textbf{18.0} & 16.5 & 14.4 & 12.7 \\
DS2 (F1) & 16.9 & \textbf{18.8} & 15.8 & 13.8 & 10.5 \\
DS3 (F1) & 37.1 & \textbf{40.7} & 37.6 & 35.7 & 31.8 \\
DS4 (F1) & \textbf{62.3} & 58.9 & 55.5 & 50.8 & 46.6 \\
DS5 (F1) & 52.9 & \textbf{85.1} & 81.6 & 76.0 & 75.5 \\
\bottomrule
\multicolumn{6}{l}{\scriptsize * Base Case Configuration}
\end{tabular}
\end{table}

\subsubsection{Dropout}

\begin{table}[H]
\centering
\caption{Average Performance for Dropout Rates.}
\label{tab:dropout_avg_performance}
\small
\setlength{\tabcolsep}{22pt}
\renewcommand{\arraystretch}{1.4}
\begin{tabular}{@{}lcccc@{}}
\toprule
\textbf{Metrics (Average)} & \textbf{0.0*} & \textbf{0.2} & \textbf{0.3} & \textbf{0.5} \\
\midrule
Macro F1 & \textbf{52.9\%} & 52.5\% & 52.7\% & 52.1\% \\
Standard Top-1 & 71.7\% & \textbf{71.8\%} & \textbf{71.8\%} & 71.4\% \\
Ordinal Top-1 & \textbf{86.2\%} & \textbf{86.2\%} & \textbf{86.2\%} & 85.8\% \\
Params (M) & 58.77 & 61.28 & 61.28 & 61.28 \\
GFLOPs & 8.98 & 9.85 & 9.85 & 9.85 \\
Inference Time (ms) & 11.16 & 11.71 & 11.69 & 11.63 \\
\midrule
DS0 (F1) & \textbf{92.4} & 84.6 & 84.5 & 84.3 \\
DS1 (F1) & 17.8 & 20.1 & \textbf{21.4} & 19.8 \\
DS2 (F1) & 16.9 & 19.6 & \textbf{19.9} & 19.0 \\
DS3 (F1) & 37.1 & 41.6 & \textbf{42.5} & 41.8 \\
DS4 (F1) & \textbf{62.3} & 62.0 & 61.2 & 60.9 \\
DS5 (F1) & 52.9 & \textbf{87.3} & 86.6 & 86.5 \\
\bottomrule
\multicolumn{5}{l}{\scriptsize * Base Case Configuration}
\end{tabular}
\end{table}

\subsubsection{Activation Function}

\begin{table}[H]
\centering
\caption{Average Performance for Activation Function Configurations.}
\label{tab:activation_avg_performance}
\small
\renewcommand{\arraystretch}{1.4}
\setlength{\tabcolsep}{8pt}
\begin{tabular}{@{}lcccccc@{}}
\toprule
\textbf{Metrics (Average)} & \textbf{Default*} & \textbf{ReLU} & \textbf{Leaky ReLU} & \textbf{ELU} & \textbf{SiLU} & \textbf{GELU} \\
\midrule
Macro F1 & \textbf{52.9\%} & 50.5\% & 49.6\% & 44.5\% & 46.5\% & 48.7\% \\
Standard Top-1 & \textbf{71.7\%} & 70.5\% & 70.3\% & 69.2\% & 69.8\% & 70.5\% \\
Ordinal Top-1 & \textbf{86.2\%} & 85.2\% & 84.5\% & 83.2\% & 83.8\% & 84.4\% \\
Params (M) & 58.77 & 61.28 & 61.28 & 61.28 & 61.28 & 61.28 \\
GFLOPs & 8.98 & 9.85 & 9.85 & 9.85 & 9.85 & 9.85 \\
Inference Time (ms) & 11.16 & 11.60 & 11.60 & 11.61 & 11.52 & 11.57 \\
\midrule
DS0 (F1) & \textbf{92.4} & 83.8 & 83.8 & 83.0 & 83.1 & 83.8 \\
DS1 (F1) & \textbf{17.8} & 17.7 & 16.3 & 14.9 & 14.2 & 15.7 \\
DS2 (F1) & \textbf{16.9} & 16.8 & 16.3 & 11.3 & 13.5 & 16.1 \\
DS3 (F1) & 37.1 & \textbf{39.6} & 37.6 & 32.8 & 35.2 & 37.7 \\
DS4 (F1) & \textbf{62.3} & 58.6 & 57.9 & 48.4 & 51.8 & 56.8 \\
DS5 (F1) & 52.9 & \textbf{86.4} & 85.9 & 76.4 & 81.3 & 82.0 \\
\bottomrule
\multicolumn{7}{l}{\scriptsize * Base Case Configuration (varies by model family)}
\end{tabular}
\end{table}


\begin{thebibliography}{00}
\bibitem{b1} Vinodkumar Devarajan, Integrated AI-ML framework for disaster lifecycle management: From prediction to recovery, World Journal of Advanced Research and Reviews 26 (2025) 585–593. https://doi.org/10.30574/wjarr.2025.26.2.1630.

\bibitem{b2} R. Gupta, M. Shah, RescueNet: Joint Building Segmentation and Damage Assessment from Satellite Imagery, in: 2020 25th International Conference on Pattern Recognition (ICPR), IEEE, 2021: pp. 4405–4411. https://doi.org/10.1109/ICPR48806.2021.9412295.

\bibitem{b3} A.M. Braik, M. Koliou, Post-tornado automated building damage evaluation and recovery prediction by integrating remote sensing, deep learning, and restoration models, Sustain. Cities Soc. 123 (2025) 106286. https://doi.org/10.1016/j.scs.2025.106286.
\bibitem{b4} J.W. van de Lindt, W. “Lisa” Wang, B. Johnston, P.S. Crawford, G. Yan, T. Dao, T. Do, K. Skakel, M. Harati, T. Nguyen, R. Umeike, S. Croope, Social Susceptibility–Driven Longitudinal Tornado Reconnaissance Methodology: 2021 Midwest Quad-State Tornado Outbreak, ASCE OPEN: Multidisciplinary Journal of Civil Engineering 3 (2025). https://doi.org/10.1061/AOMJAH.AOENG-0065.
\bibitem{b5} M. Imran, U. Qazi, F. Ofli, S. Peterson, F. Alam, AI for Disaster Rapid Damage Assessment from Microblogs, Proceedings of the AAAI Conference on Artificial Intelligence 36 (2022) 12517–12523. https://doi.org/10.1609/aaai.v36i11.21521.
\bibitem{b6} A. Jones, J. Kuehnert, P. Fraccaro, O. Meuriot, T. Ishikawa, B. Edwards, N. Stoyanov, S.L. Remy, K. Weldemariam, S. Assefa, AI for climate impacts: applications in flood risk, NPJ Clim. Atmos. Sci. 6 (2023) 63. https://doi.org/10.1038/s41612-023-00388-1. 

\bibitem{b7} M.S. Hackathon, Leveraging AI for Natural Disaster Management : Takeaways From The Moroccan Earthquake, (2023). http://arxiv.org/abs/2311.08999.

\bibitem{b8} Y. Senarath, R. Pandey, S. Peterson, H. Purohit, Citizen-Helper System for Human-Centered AI Use in Disaster Management, in: A. Singh (Ed.), International Handbook of Disaster Research, Springer Nature Singapore, Singapore, 2023: pp. 477–497. https://doi.org/10.1007/978-981-19-8388-7\_34.

\bibitem{b9} W. “Lisa” Wang, J.W. van de Lindt, B. Johnston, P.S. Crawford, G. Yan, T. Dao, T. Do, K. Skakel, M. Harati, T. Nguyen, R. Umeike, S. Croope, A. R. Barbosa, Application of Multidisciplinary Community Resilience Modeling to Reduce Disaster Risk: Building Back Better, Journal of Performance of Constructed Facilities 38 (2024). https://doi.org/10.1061/JPCFEV.CFENG-4650.
\bibitem{b10} B. Johnston, L. Wang, J.W. van de Lindt, M. Harati, K. Skakel, S. Crawford, T. Dao, C. Robinson, G. Yan, T. Do, J. Loerzel, A. Barbosa, S. Croope, Interdisciplinary data collection for empirical community-level recovery modelling, in: 2024: pp. 1260–1267. https://doi.org/10.2749/manchester.2024.1260.
\bibitem{b11} C. Yu, B. Hu, X. Cheng, G. Yin, Z. Wang, Remote sensing building damage assessment with a multihead neighbourhood attention transformer, Int. J. Remote Sens. 44 (2023) 5069–5100. https://doi.org/10.1080/01431161.2023.2242590.
\bibitem{b12} F. Zhao, C. Zhang, Building Damage Evaluation from Satellite Imagery using Deep Learning, in: 2020 IEEE 21st International Conference on Information Reuse and Integration for Data Science (IRI), IEEE, 2020: pp. 82–89. https://doi.org/10.1109/IRI49571.2020.00020.
\bibitem{b13} F. Xiong, H. Wen, C. Zhang, C. Song, X. Zhou, Semantic segmentation recognition model for tornado-induced building damage based on satellite images, Journal of Building Engineering 61 (2022) 105321. https://doi.org/10.1016/j.jobe.2022.105321.
\bibitem{b14} R. Umeike, T. Dao, S. Crawford, Accelerating Post-Tornado Disaster Assessment Using Advanced Deep Learning Models, in: 2024 IEEE MetroCon, IEEE, 2024: pp. 1–3. https://doi.org/10.1109/MetroCon62511.2024.10883935.
\bibitem{b15} K. Ahn, S. Han, S. Park, J. Kim, S. Park, M. Cha, Generalizable Disaster Damage Assessment via Change Detection with Vision Foundation Model, Proceedings of the AAAI Conference on Artificial Intelligence 39 (2025) 27784–27792. https://doi.org/10.1609/aaai.v39i27.34994.
\bibitem{b16} E. Weber, N. Marzo, D.P. Papadopoulos, A. Biswas, A. Lapedriza, F. Ofli, M. Imran, A. Torralba, Detecting Natural Disasters, Damage, and Incidents in the Wild, in: 2020: pp. 331–350. https://doi.org/10.1007/978-3-030-58529-7\_20.
\bibitem{b17} S. Sonang, Hybrid CNN Approach for Post-Disaster Building Damage Classification Using Satellite Imagery, Journal of Applied Data Sciences 6 (2025) 2824–2837. https://doi.org/10.47738/jads.v6i4.931.

\bibitem{b18} D.P. Kingma, J. Ba, Adam: A Method for Stochastic Optimization, in: Y. Bengio, Y. LeCun (Eds.), 3rd International Conference on Learning Representations, ICLR 2015, San Diego, CA, USA, May 7-9, 2015, Conference Track Proceedings, 2015. http://arxiv.org/abs/1412.6980.

\bibitem{b19} T. Dalvi, A. Bhatt, A Deep Learning-Based Approach for Automatic Detection of Hurricane Damage using Satellite Images, in: 2023 IEEE 8th International Conference for Convergence in Technology (I2CT), IEEE, 2023: pp. 1–5. https://doi.org/10.1109/I2CT57861.2023.10126276.

\bibitem{b20} Y.-S. Chu, H.-C. Wei, Post-Disaster Affected Area Segmentation with a Vision Transformer (ViT)-based EVAP Model using Sentinel-2 and Formosat-5 Imagery, in: 2025.

\bibitem{b21} Z. Zheng, Y. Zhong, L. Zhang, M. Burke, D.B. Lobell, S. Ermon, Towards transferable building damage assessment via unsupervised single-temporal change adaptation, Remote Sens. Environ. 315 (2024) 114416. https://doi.org/10.1016/j.rse.2024.114416.

\bibitem{b22} J. Chang, Y. Cen, G. Cen, Asymmetric Network Combining CNN and Transformer for Building Extraction from Remote Sensing Images, Sensors 24 (2024) 6198. https://doi.org/10.3390/s24196198.

\bibitem{b23} J. Deng, W. Dong, R. Socher, L.-J. Li, Kai Li, Li Fei-Fei, ImageNet: A large-scale hierarchical image database, in: 2009 IEEE Conference on Computer Vision and Pattern Recognition, IEEE, 2009: pp. 248–255. https://doi.org/10.1109/CVPR.2009.5206848.

\bibitem{b24} B. Zhou, A. Lapedriza, A. Khosla, A. Oliva, A. Torralba, Places: A 10 Million Image Database for Scene Recognition, IEEE Trans. Pattern Anal. Mach. Intell. 40 (2018) 1452–1464. https://doi.org/10.1109/TPAMI.2017.2723009.

\bibitem{b25} U. Lagap, S. Ghaffarian, Enhancing Post-Disaster Damage Detection and Recovery Monitoring by Addressing Class Imbalance in Satellite Imagery Using Enhanced Super-Resolution GANs (ESRGAN), The International Archives of the Photogrammetry, Remote Sensing and Spatial Information Sciences XLVIII-G-2025 (2025) 853–860. https://doi.org/10.5194/isprs-archives-XLVIII-G-2025-853-2025.

\bibitem{b26} K. Dunphy, M.N. Fekri, K. Grolinger, A. Sadhu, Data Augmentation for Deep-Learning-Based Multiclass Structural Damage Detection Using Limited Information, Sensors 22 (2022) 6193. https://doi.org/10.3390/s22166193.

\bibitem{b27} Z. Hong, H. Zhong, H. Pan, J. Liu, R. Zhou, Y. Zhang, Y. Han, J. Wang, S. Yang, C. Zhong, Classification of Building Damage Using a Novel Convolutional Neural Network Based on Post-Disaster Aerial Images, Sensors 22 (2022) 5920. https://doi.org/10.3390/s22155920.

\bibitem{b28} X. Chen, Using satellite imagery to automate building damage assessment: A case study of the xbd dataset, (2021).

\bibitem{b29} A.E. Yilmaz, H. Demirhan, Weighted kappa measures for ordinal multi-class classification performance, Appl. Soft Comput. 134 (2023) 110020. https://doi.org/10.1016/j.asoc.2023.110020.

\bibitem{b30} A. Chakravarty, M. Brenchley, T. Breakspear, I. Lewin, Y. Huang, Enhancing Marker Scoring Accuracy through Ordinal Confidence Modelling in Educational Assessments, in: Proceedings of the 63rd Annual Meeting of the Association for Computational Linguistics (Volume 6: Industry Track), Association for Computational Linguistics, Stroudsburg, PA, USA, 2025: pp. 1498–1507. https://doi.org/10.18653/v1/2025.acl-industry.106.

\bibitem{b31} A. Graettinger, C. Ramseyer, S. Freyne, D. Prevatt, L. Myers, T. Dao, R. Floyd, L. Holliday, D. Agdas, F. Haan, J. Richardson, R. Gupta, R. Emerson, C. Alfano, Tornado damage assessment in the aftermath of the May 20th 2013 Moore Oklahoma tornado, 2014.

\bibitem{b32} A. Krizhevsky, I. Sutskever, G.E. Hinton, ImageNet classification with deep convolutional neural networks, Commun. ACM 60 (2017) 84–90. https://doi.org/10.1145/3065386.

\bibitem{b33} K. Simonyan, A. Zisserman, Very deep convolutional networks for large-scale image recognition, in: 3rd International Conference on Learning Representations (ICLR 2015), Computational and Biological Learning Society, 2015: pp. 1–14.

\bibitem{b34} K. He, X. Zhang, S. Ren, J. Sun, Deep Residual Learning for Image Recognition, in: 2016 IEEE Conference on Computer Vision and Pattern Recognition (CVPR), IEEE, 2016: pp. 770–778. https://doi.org/10.1109/CVPR.2016.90.

\bibitem{b35} S. Xie, R. Girshick, P. Dollar, Z. Tu, K. He, Aggregated Residual Transformations for Deep Neural Networks, in: 2017 IEEE Conference on Computer Vision and Pattern Recognition (CVPR), IEEE, 2017: pp. 5987–5995. https://doi.org/10.1109/CVPR.2017.634.

\bibitem{b36} S. Zagoruyko, N. Komodakis, Wide Residual Networks, ArXiv abs/1605.07146 (2016). https://api.semanticscholar.org/CorpusID:15276198.

\bibitem{b37} G. Huang, Z. Liu, L. Van Der Maaten, K.Q. Weinberger, Densely Connected Convolutional Networks, in: 2017 IEEE Conference on Computer Vision and Pattern Recognition (CVPR), IEEE, 2017: pp. 2261–2269. https://doi.org/10.1109/CVPR.2017.243.

\bibitem{b38} M. Tan, Q. Le, EfficientNet: Rethinking Model Scaling for Convolutional Neural Networks, in: K. Chaudhuri, R. Salakhutdinov (Eds.), Proceedings of the 36th International Conference on Machine Learning, PMLR, 2019: pp. 6105–6114. https://proceedings.mlr.press/v97/tan19a.html.

\bibitem{b39} A.G. Howard, M. Zhu, B. Chen, D. Kalenichenko, W. Wang, T. Weyand, M. Andreetto, H. Adam, MobileNets: Efficient Convolutional Neural Networks for Mobile Vision Applications., CoRR abs/1704.04861 (2017). http://dblp.uni-trier.de/db/journals/corr/corr1704.html\#HowardZCKWWAA17.

\bibitem{b40} X. Zhang, X. Zhou, M. Lin, J. Sun, ShuffleNet: An Extremely Efficient Convolutional Neural Network for Mobile Devices, in: 2018 IEEE/CVF Conference on Computer Vision and Pattern Recognition, IEEE, 2018: pp. 6848–6856. https://doi.org/10.1109/CVPR.2018.00716.

\bibitem{b41} F.N. Iandola, M.W. Moskewicz, K. Ashraf, S. Han, W.J. Dally, K. Keutzer, SqueezeNet: AlexNet-level accuracy with 50x fewer parameters and <1MB model size, ArXiv abs/1602.07360 (2016). https://api.semanticscholar.org/CorpusID:14136028.

\bibitem{b42} C. Szegedy, Wei Liu, Yangqing Jia, P. Sermanet, S. Reed, D. Anguelov, D. Erhan, V. Vanhoucke, A. Rabinovich, Going deeper with convolutions, in: 2015 IEEE Conference on Computer Vision and Pattern Recognition (CVPR), IEEE, 2015: pp. 1–9. https://doi.org/10.1109/CVPR.2015.7298594.

\bibitem{b43} I. Radosavovic, R.P. Kosaraju, R. Girshick, K. He, P. Dollar, Designing Network Design Spaces, in: 2020 IEEE/CVF Conference on Computer Vision and Pattern Recognition (CVPR), IEEE, 2020: pp. 10425–10433. https://doi.org/10.1109/CVPR42600.2020.01044.

\bibitem{b44} Z. Liu, H. Mao, C.-Y. Wu, C. Feichtenhofer, T. Darrell, S. Xie, A ConvNet for the 2020s, in: 2022 IEEE/CVF Conference on Computer Vision and Pattern Recognition (CVPR), IEEE, 2022: pp. 11966–11976. https://doi.org/10.1109/CVPR52688.2022.01167.

\bibitem{b45} A. Dosovitskiy, L. Beyer, A. Kolesnikov, D. Weissenborn, X. Zhai, T. Unterthiner, M. Dehghani, M. Minderer, G. Heigold, S. Gelly, J. Uszkoreit, N. Houlsby, An Image is Worth 16x16 Words: Transformers for Image Recognition at Scale, in: International Conference on Learning Representations, 2021. https://openreview.net/forum?id=YicbFdNTTy.

\bibitem{b46} Z. Liu, Y. Lin, Y. Cao, H. Hu, Y. Wei, Z. Zhang, S. Lin, B. Guo, Swin Transformer: Hierarchical Vision Transformer using Shifted Windows, in: 2021 IEEE/CVF International Conference on Computer Vision (ICCV), IEEE, 2021: pp. 9992–10002. https://doi.org/10.1109/ICCV48922.2021.00986.

\bibitem{b47} Z. Tu, H. Talebi, H. Zhang, F. Yang, P. Milanfar, A. Bovik, Y. Li, MaxViT: Multi-axis Vision Transformer, in: 2022: pp. 459–479. https://doi.org/10.1007/978-3-031-20053-3\_27


\end{thebibliography}
\end{document}